\documentclass{article}

\usepackage[preprint]{neurips_2026}

\usepackage[utf8]{inputenc} 
\usepackage[T1]{fontenc}    
\usepackage{hyperref}       
\hypersetup{hidelinks}
\usepackage{url}            
\usepackage{booktabs}       
\usepackage{amsfonts}       
\usepackage{nicefrac}       
\usepackage{microtype}      
\usepackage{xcolor}         

\usepackage{amsmath}
\usepackage{mathtools}
\usepackage{amsthm}
\usepackage{graphicx}
\usepackage{algorithm}
\usepackage{multirow}
\usepackage{makecell}
\usepackage{listings}
\usepackage{caption}
\usepackage{amssymb}
\usepackage{subcaption}

\title{CoANeRV: Coordinate-Aware Token-Space \\ Neural Video Representation}

\author{%
Jialong Guo \quad
Ke Liu \quad
Mengxuan Li \quad
Jiajun Bu \quad
Haishuai Wang\thanks{Corresponding author: Haishuai Wang.}\\
Zhejiang Key Laboratory of Accessible Perception and Intelligent Systems\\
College of Computer Science, Zhejiang University, China\\
\texttt{\{jialongguo,keliu99,limx97,bjj,haishuai.wang\}@zju.edu.cn}
}

\begin{document}

\maketitle

\begin{abstract}

Neural representations for videos (NeRV) have shown strong reconstruction fidelity by storing video-specific information in network weights. However, existing formulations typically require either costly per-video optimization or video-specific weight generation, making it difficult to scale to efficient amortized video representation.
We propose \textbf{CoANeRV}, a coordinate-aware token-space framework that adapts the broader token-conditioned neural-field paradigm to amortized video representation. CoANeRV forms compact video tokens in one feed-forward pass and uses a shared coordinate-conditioned decoder to reconstruct continuous spatio-temporal queries, avoiding per-video decoder optimization or generation while retaining coordinate-level reconstruction flexibility.
To make token-space reconstruction effective, CoANeRV introduces a coordinate-aware decoding architecture that aligns spatio-temporal queries with video tokens through axis-adaptive positional encoding and temperature-modulated cross-attention. Block-wise coordinate querying further reduces peak attention memory, making high-resolution reconstruction practical.
Experiments on diverse video datasets show that CoANeRV consistently improves reconstruction quality over prior feed-forward NeRV and INR baselines, reduces peak memory compared with attention-based coordinate decoders, and provides efficient amortized encoding without per-video optimization. These results support the proposed video-specific combination of feed-forward token formation, spatio-temporal coordinate retrieval, and memory-bounded dense querying. The code is available at \url{https://github.com/jialong2023/CoANeRV}.

\end{abstract}

\vspace{-20pt}
\section{Introduction}
\label{sec:intro}
\vspace{-5pt}

Efficient video representation remains a central challenge in computer vision, as the high-dimensional nature of video data imposes substantial computational and storage demands.
The rapid growth of video content across streaming platforms, autonomous systems, and multimedia applications has heightened the need for representations that capture both spatial and temporal information while maintaining favorable storage, encoding, and reconstruction trade-offs.
Traditional video compression standards such as H.264/AVC~\cite{wiegand2003overview} and HEVC~\cite{sullivan2012overview} are highly optimized for rate--distortion performance, but their block-based and entropy-coded pipelines offer limited flexibility as continuous, task-adaptive neural representations.

Neural video representations provide an alternative to conventional codecs by modeling visual signals with neural functions~\cite{li2022enerv,kim2022scalable}.
NeRV-style methods~\cite{chen2021nerv,chen2022cnerv,chen2023hnerv} encode a video into network parameters and reconstruct frames from temporal indices, achieving compact storage and high reconstruction fidelity.
However, this weight-space formulation couples video identity with decoder parameters: each new video typically requires either per-video optimization or the generation of video-specific weights.

Recent feed-forward video methods amortize weight generation~\cite{guo2025metanerv,chen2024fast}, while token-conditioned neural fields already establish per-instance tokens, shared coordinate decoders, and cross-attention for other continuous signals~\cite{lee2023laginr,haider2025tokenized,urbano2026hierarchy}. We ask how this broader paradigm can support amortized complete-clip encoding, anisotropic $(x,y,t)$ queries, and memory-bounded dense video reconstruction without generating or optimizing a decoder per video.

\begin{figure*}[t]
    \centering
    \vspace{-10pt}
    \includegraphics[width=0.99\textwidth]{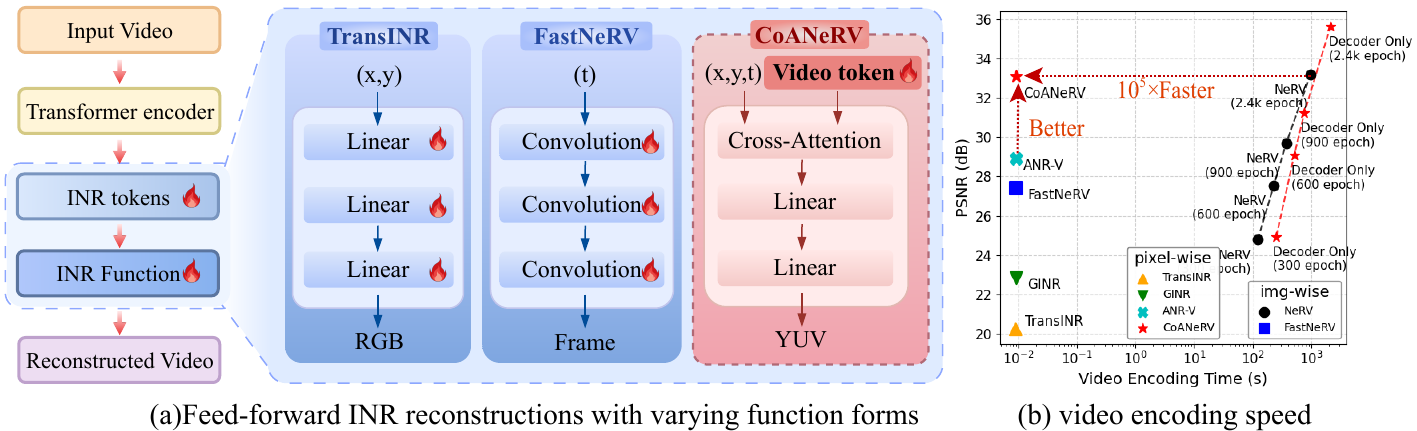}
    \caption{
    \textbf{Framework comparison and encoding speed.}
    (a) Representative weight-space video approaches store instance-specific information in optimized, generated, or modulated decoder parameters, whereas CoANeRV stores video-specific information in compact tokens and reconstructs pixels with a shared coordinate-conditioned decoder.
    (b) CoANeRV enables feed-forward token inference, while the decoder-only variant freezes the shared decoder and optimizes only video tokens to isolate the cost of per-video token fitting.
    }
    \label{fig:framework1}
    \vspace{-14pt}
\end{figure*}

We propose \textbf{CoANeRV}, an amortized video realization of the token-conditioned neural-field paradigm. A feed-forward encoder forms fixed-length memory from complete-clip patches, while a shared decoder uses anisotropic spatio-temporal queries and exact memory-bounded dense execution. Our claim is this video-specific coupling, not tokens, shared coordinate decoders, or cross-attention in isolation.

Figure~\ref{fig:framework1} illustrates the CoANeRV pipeline and contrasts it with existing feed-forward and per-video optimized approaches.
Accordingly, CoANeRV should be understood as a token-space coordinate-based neural video representation rather than a pure weight-space NeRV model: video-specific content is stored in latent tokens, while the shared decoder provides a persistent coordinate-to-signal mapping.
Our contributions are summarized as follows:

\begin{itemize}
    \item
    We adapt token-conditioned neural fields to amortized video representation through a feed-forward complete-clip tokenizer and token former. The resulting fixed-length video memory removes the need to optimize or generate a decoder for each video while retaining coordinate-level reconstruction flexibility.

    \item
    We design a shared coordinate-conditioned decoder that reconstructs videos by aligning spatio-temporal coordinate queries with video tokens. Axis-adaptive positional encoding and temperature-modulated cross-attention improve coordinate--token interaction, while block-wise coordinate querying reduces peak attention memory for high-resolution reconstruction.

    \item
    Experiments on diverse video datasets show that CoANeRV improves reconstruction quality over prior feed-forward NeRV and INR baselines, reduces peak memory compared with attention-based coordinate decoders, and enables efficient amortized encoding without per-video optimization.
\end{itemize}

\vspace{-7pt}

\vspace{-11pt}
\section{Related Work}
\vspace{-5pt}

\noindent\textbf{Implicit Neural Representations.}
Implicit neural representations provide a compact approach to representing visual signals as continuous neural functions~\cite{dupont2021coin, chen2022videoinr}.
Foundational coordinate-based methods~\cite{tancik2020fourier, sitzmann2020implicit} use multilayer perceptrons to map spatial or spatiotemporal coordinates to signal values, demonstrating strong performance in applications such as novel view synthesis~\cite{mildenhall2021nerf} and image super-resolution~\cite{chen2021learning}.
Building on these approaches, NeRV~\cite{chen2021nerv} introduced frame-wise implicit representations, generating entire frames from temporal indices via convolutional networks.
Subsequent works improved reconstruction quality through various strategies, including E-NeRV~\cite{li2022enerv} with spatial-temporal decomposition and HNeRV~\cite{chen2023hnerv} with hybrid autoencoder designs.
Despite these advances, coordinate-based INRs often achieve high reconstruction fidelity in specific scenarios~\cite{chen2022videoinr, kim2022scalable}, highlighting a trade-off between feed-forward efficiency and coordinate-level reconstruction quality.

\vspace{-3pt}
\noindent\textbf{Generalizable and Hypernetwork-Based INR Representations.}
Hypernetworks~\cite{ha2016hypernetworks} generate adaptive model parameters conditioned on input data and have been used to synthesize instance-specific neural representations~\cite{park2019deepsdf, mescheder2019occupancy}.
Recent generalizable INR methods amortize instance-specific representation learning through transformer encoders, hypernetworks, or feed-forward weight-generation mechanisms, including TransINR~\cite{chen2022transinr}, GINR~\cite{kim2022generalizable}, FastNeRV~\cite{chen2024fast}, and ANR~\cite{zhang2024attention}.
The closest precedents are token-conditioned neural fields. LAG-INR~\cite{lee2023laginr} already combines feed-forward instance tokens, coordinate cross-attention, and a shared decoder for 2D coordinates/6D rays. Tokenized Neural Fields~\cite{haider2025tokenized}, a NeurIPS 2025 workshop paper, instead fits new-instance tokens by gradient optimization. The concurrent June 2026 preprint LH-NeF~\cite{urbano2026hierarchy} forms hierarchical locality-aware tokens with group routing, Gaussian cross-attention, and feature-wise linear modulation over general metric domains. Thus, tokens, shared coordinate decoders, and cross-attention are prior art.

CoANeRV contributes their video-specific coupling: feed-forward complete-clip memory, anisotropic $(x,y,t)$ encoding, temperature-controlled retrieval, and exact tiled access to the full token bank. Our matched ANR-V comparison isolates the video retrieval rule; Appendix~\ref{app:related_token_fields} provides the cross-paper technical comparison without numerical claims across incompatible tasks.

\vspace{-3pt}
\noindent\textbf{Video Codecs.}
Traditional video codecs such as H.265/HM~\cite{HM}, H.266/VTM~\cite{VTM}, and ECM~\cite{ECM} remain strong compression baselines with highly optimized rate--distortion performance, but their entropy-coded bitstreams are not designed as continuous neural representations.
Learning-based neural video codecs~\cite{Rippel_2019_ICCV, agustsson2020scale, maiya2023nirvana} have improved learned rate--distortion optimization, although decoding latency and implementation complexity remain important practical considerations.
Neural codecs have evolved along two directions: the DCVC family~\cite{li2021deep, sheng2022temporal, li2022hybrid, li2023neural, li2024neural, jia2025towards} prioritizes compression efficiency and real-time performance using conditional coding and entropy modeling, while representation-oriented methods such as NeRV~\cite{chen2021nerv} and DNeRV~\cite{he2023dnerv} focus on reconstruction fidelity and neural signal representation.
CoANeRV follows this representation-first paradigm, introducing coordinate-aware decoding and block query processing to improve reconstruction quality while reducing peak attention memory.

\vspace{-3pt}
\noindent\textbf{Temperature-Modulated Attention.}
Temperature scaling is commonly used to control attention sharpness in sequence and vision models~\cite{lin2018learning,zhang2021attention,zhou2023learning}.
In CoANeRV, it is not treated as a standalone attention contribution; instead, it serves as a simple mechanism for regulating coordinate-token retrieval within the proposed token-space video representation framework.
Prior work in natural language processing has demonstrated its utility for improving summarization~\cite{zhang2021attention} and machine translation~\cite{lin2018learning}.
In computer vision, temperature-scaled attention has been applied to tasks such as image inpainting~\cite{zhou2023learning} to better exploit contextual information.
CoANeRV applies \textit{temperature-modulated cross-attention} to coordinate-based video reconstruction, adjusting attention distributions between coordinate queries and instance-specific video tokens.
This mechanism improves coordinate-to-token alignment and reconstruction fidelity; combined with block query processing, it enables memory-efficient coordinate-aware reconstruction.

\textbf{Positioning of CoANeRV.}
CoANeRV is not intended to replace modern neural video codecs such as DCVC-style systems, which are optimized for entropy coding, motion compensation, and real-time rate--distortion performance. Instead, CoANeRV targets representation-oriented video reconstruction, where fast amortized encoding, shared decoding parameters, and coordinate-query access are desirable. We therefore compare primarily with feed-forward NeRV-style and coordinate-based neural representation methods, while reporting codec comparisons as an auxiliary analysis.

\vspace{-15pt}
\section{Method}
\label{sec:method}
\vspace{-5pt}

\subsection{Problem Statement}
\vspace{-5pt}

Let \(V \in \mathbb{R}^{T \times C \times H \times W}\) denote a video sequence.
Conventional NeRV-style representations encode a video in the parameters of a neural decoder \(f_{\theta_V}\), which maps a temporal index to a reconstructed frame:
\begin{equation}
\hat{V}_t = f_{\theta_V}(t), \quad t \in \{0,\ldots,T-1\}.
\end{equation}
Here the video-specific information is stored in the instance-specific parameter set \(\theta_V\), which is obtained either by per-video optimization or by a feed-forward weight-generation network.

CoANeRV considers a token-space alternative.
Instead of assigning each video an instance-specific decoder, we represent video-specific content with compact video tokens \(\mathbf{T}_v\) and use a decoder \(D_\psi\) shared across all videos:
\begin{equation}
\mathbf{T}_v = E_\phi(V), \qquad
\hat{V}(q) = D_\psi(\gamma(q), \mathbf{T}_v), \quad q=(x,y,t).
\end{equation}
The goal is to learn \(E_\phi\) and \(D_\psi\) such that \(\mathbf{T}_v\) captures video-specific content, while the persistent decoder provides a shared coordinate-to-signal mapping.
This formulation shifts video-specificity from weight space to token space.

\begin{figure*}[!t]
\vspace{-20pt}
\centering
\includegraphics[width=0.99\textwidth]{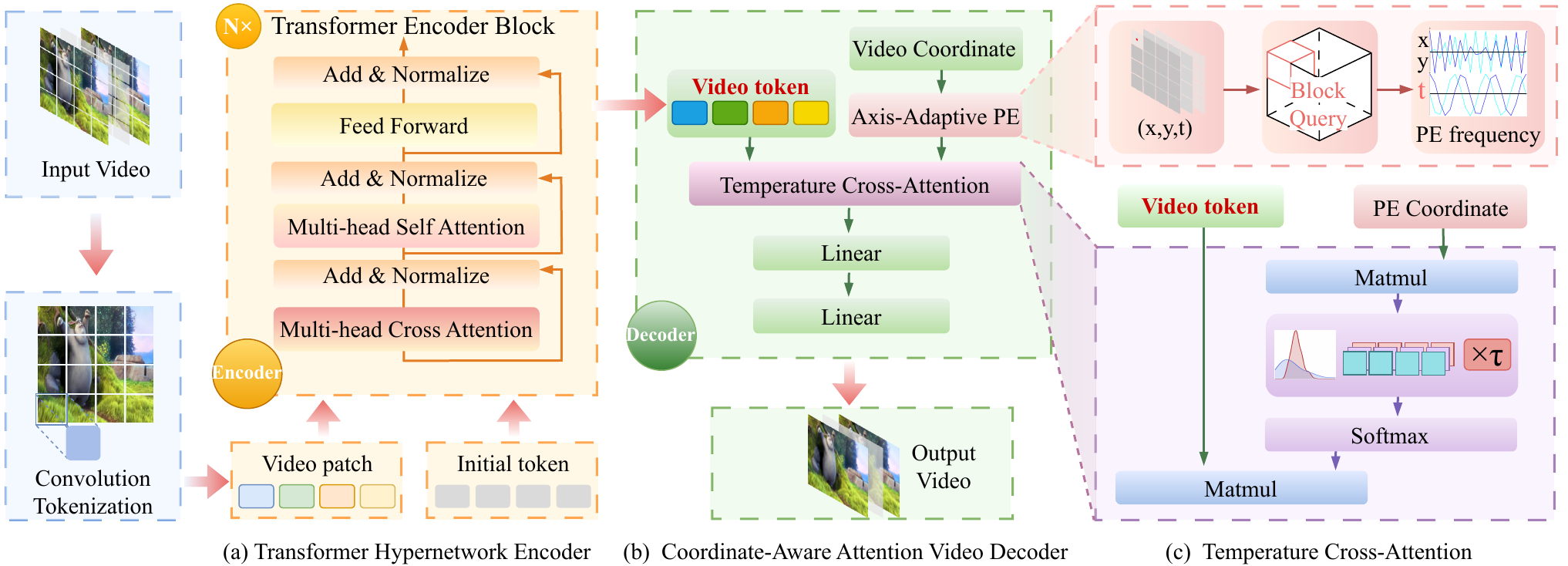}
\caption{
\textbf{CoANeRV framework.}
CoANeRV encodes each video through convolutional patch tokenization and a six-layer token former, shown as the ``Transformer Encoder Block'' in the schematic, and reconstructs pixels by querying the resulting compact video tokens with continuous spatio-temporal coordinates. The decoder is shared across videos, so video-specificity is stored in token space rather than decoder weight space.
}
\label{fig:framework2}
\vspace{-20pt}
\end{figure*}

\vspace{-5pt}
\subsection{Overall Workflow}
\vspace{-5pt}

As shown in Figure~\ref{fig:framework2}, CoANeRV consists of a video token encoder and a shared coordinate-conditioned decoder.
Given an input video \(V\), the encoder produces a compact token set
\begin{equation}
\mathbf{T}_v = E_\phi(V), \qquad \mathbf{T}_v \in \mathbb{R}^{N \times d},
\end{equation}
where \(N\) is the number of video tokens and \(d\) is the token dimension.
For any normalized spatio-temporal coordinate \(q=(x,y,t)\), the decoder reconstructs the corresponding signal value by conditioning on both the coordinate embedding and the video tokens:
\begin{equation}
\hat{V}(q) = D_\psi(\gamma(q), \mathbf{T}_v).
\end{equation}

For training, we use coordinates from the normalized discrete grid implemented by the coordinate generator:
\begin{equation}
\Omega = \left\{
\left(\frac{x}{W-1}, \frac{y}{H-1}, \frac{t}{T}\right)
\,\middle|\,
0 \leq x < W,\; 0 \leq y < H,\; 0 \leq t < T
\right\}.
\end{equation}
Thus, the spatial coordinates include both endpoints of $[0,1]$, while the temporal coordinates follow the half-open grid $\{0,1/T,\ldots,(T-1)/T\}$ used by the implementation.
The model is trained end-to-end with the reconstruction loss
\begin{equation}
\mathcal{L} =
\frac{1}{|\Omega|}
\sum_{q\in\Omega}
\left\|V(q)-\hat{V}(q)\right\|_2^2.
\end{equation}

At test time, a video is encoded by a single forward pass \(\mathbf{T}_v=E_\phi(V)\).
No decoder weights are generated or optimized for individual videos; all video-specific information is carried by \(\mathbf{T}_v\).

For a collection of $M_v$ videos and a decoder with $p_{\mathrm{dec}}$ parameters, storing one decoder per video scales as $\mathcal{O}(M_vp_{\mathrm{dec}})$, whereas CoANeRV stores the shared decoder once and one $N\times d$ token bank per video, scaling as $\mathcal{O}(p_{\mathrm{dec}}+M_vNd)$. This is representation-storage scaling, excluding shared encoder parameters and any codec bitstream machinery; query-time complexity is stated with the block-wise decoder below.

\vspace{-10pt}
\subsection{Convolutional Video Tokenization and Token Former}
\vspace{-5pt}

Token-conditioned shared decoders have precedents~\cite{lee2023laginr,haider2025tokenized,urbano2026hierarchy}. CoANeRV specializes them to amortized video representation: $E_\phi$ maps complete-clip patches to fixed-length video memory in one pass, without fitting tokens at test time or generating decoder weights.

\textbf{Convolutional Tokenization.}
We first map the input video into spatiotemporal patch features $\mathbf{P} \in \mathbb{R}^{N_p \times D}$ using a learnable convolutional tokenizer. Here \(N_p\) denotes the number of initial patch tokens, and $D$ is the intermediate token-former width; in the main configuration, $D=768$, whereas the final stored video-token dimension is $d=72$.
Compared with explicitly unfolding patches followed by a linear projection, the strided convolutional tokenizer provides an efficient and implementation-friendly patch projection while preserving local patch structure before token compression.

\textbf{Video Token Formation.}
To obtain a compact representation, we initialize $N$ learned query tokens $\mathbf{R}^{(0)} \in \mathbb{R}^{N \times D}$, with $N \ll N_p$, and update them through $L$ token-former blocks:
\begin{equation}
\mathbf{R}^{(\ell)}=
\mathcal{F}_{\ell}\!\left(\mathbf{R}^{(\ell-1)},\mathbf{P}\right),
\qquad \ell=1,\ldots,L.
\end{equation}
Each block applies cross-attention from the learned queries to the patch features, followed by query-token self-attention and a feed-forward network. The final intermediate tokens are projected to the stored video-token dimension by a shared projection $\Pi_\phi$:
\begin{equation}
\mathbf{T}_v=\Pi_\phi\!\left(\mathbf{R}^{(L)}\right),
\qquad
\mathbf{T}_v\in\mathbb{R}^{N\times d}.
\end{equation}
In the main configuration, the optional patch-token self-encoder has depth zero, while the token former uses $L=6$ blocks with width $D=768$ and produces video tokens with $d=72$.

This design preserves local patch structure before compression and produces a compact content memory for coordinate-level reconstruction.
The compressed tokens \(\mathbf{T}_v\) do not instantiate decoder parameters; instead, they provide video-specific information that can be selectively retrieved by coordinate queries.
This separation is essential to the token-space formulation: the encoder determines what video content is stored, while the shared decoder determines how coordinates access and decode that content.

\vspace{-5pt}
\subsection{Coordinate-Aware Attention Video Decoder}
\label{subsec:decoder}
\vspace{-5pt}

The decoder reconstructs video frames by establishing direct correspondence between coordinates and video tokens through cross-attention. In contrast to hypernetwork-based approaches~\cite{chen2022transinr, chen2024fast}, which require generating layer-wise weights, our decoder operates with fixed parameters and performs reconstruction via a single attention-based mapping.

\textbf{Axis-Adaptive Positional Encoding.}
Given a normalized coordinate \(q=(x,y,t)\), the decoder first maps it to a multi-frequency positional embedding.
Rather than allocating the same number of frequencies to all axes, we use separate spatial and temporal frequency banks:
\begin{equation}
\gamma(q) =
[
\sin(\pi x\mathbf{w}_s), \cos(\pi x\mathbf{w}_s),
\sin(\pi y\mathbf{w}_s), \cos(\pi y\mathbf{w}_s),
\sin(\pi t\mathbf{w}_t), \cos(\pi t\mathbf{w}_t)
],
\end{equation}
where
\begin{equation}
\mathbf{w}_s = [\sigma_s^{i/(k_s-1)}]_{i=0}^{k_s-1},
\qquad
\mathbf{w}_t = [\sigma_t^{i/(k_t-1)}]_{i=0}^{k_t-1}.
\end{equation}
In our implementation, \(k_s\) and \(k_t\) are chosen to approximately follow \(k_s:k_t=4:1\) while satisfying \(2k_s+k_t=3k\), preserving the total embedding budget and allocating more frequencies to spatial axes.
This provides a simple anisotropic spatio-temporal prior: spatial coordinates often require richer high-frequency detail, while temporal variation is typically smoother under short-frame reconstruction.

\textbf{Temperature-Modulated Cross-Attention.}
The coordinate embedding queries the video tokens through temperature-modulated cross-attention:
\begin{equation}
\mathbf{F}_{\tau}(q) =
\mathrm{Softmax}\left(
\frac{\mathbf{Q}(q)\mathbf{K}(\mathbf{T}_v)^\top}{\tau\sqrt{d_h}}
\right)\mathbf{V}(\mathbf{T}_v),
\end{equation}
where
\[
\mathbf{Q}(q)=W_Q\gamma(q), \quad
\mathbf{K}(\mathbf{T}_v)=W_K \mathbf{T}_v, \quad
\mathbf{V}(\mathbf{T}_v)=W_V \mathbf{T}_v.
\]
Here \(d_h\) denotes the per-head attention dimension and \(\tau>0\) controls the selectivity of coordinate-token retrieval.
Smaller \(\tau\) produces sharper token selection, whereas larger \(\tau\) yields smoother token aggregation.
The attended feature is then mapped to pixel values by an MLP:
\begin{equation}
\hat{V}(q)=\mathrm{MLP}(\mathbf{F}_{\tau}(q)).
\end{equation}

\textbf{Block-Wise Coordinate Querying.}
Directly decoding all coordinates in \(\Omega\) requires processing \(|\Omega|N=THWN\) coordinate-token score pairs. The implementation partitions only the spatial grid into non-overlapping tiles of side length $s$, while retaining all $T$ temporal positions in every tile. Assuming $s$ divides both $H$ and $W$, this gives
\begin{equation}
\Omega=\bigcup_{b=1}^{B_s}\Omega_b,
\qquad
B_s=\frac{H}{s}\frac{W}{s},
\qquad
|\Omega_b|=Ts^2\equiv M.
\end{equation}
For each spatial tile, all $Ts^2$ coordinate queries attend to the same full token set \(\mathbf{T}_v\):
\begin{equation}
\hat{V}_{\Omega_b}=D_\psi(\gamma(\Omega_b), \mathbf{T}_v).
\end{equation}
Because the decoder contains no operation that couples different coordinate queries, tiled and untiled evaluation are mathematically equivalent; different fused or mixed-precision kernels may nevertheless introduce negligible floating-point variation. The spatial tiling reduces the peak attention-memory order from \(\mathcal{O}(THWN)\) to \(\mathcal{O}(Ts^2N)\), while preserving total arithmetic complexity \(\mathcal{O}(THWN)\). We use $s=64$ in the main experiments.

\textbf{Relation to Localized Attention.}
ANR~\cite{zhang2024attention} uses localized attention layers (LAL) for image-level INR reconstruction, where weak attention responses are suppressed by thresholding and renormalization:
\vspace{-5pt}
\begin{equation}
\label{eq:lal}
    \mathbf{F}_{\mathrm{LAL}} =
    \mathrm{Norm}\left(
    \mathrm{ReLU}\left(
    \mathrm{Softmax}\left(\frac{\mathbf{Q}\mathbf{K}^{\top}}{\sqrt{d_h}}\right) - m
    \right)
    \right)\mathbf{V}.
\end{equation}
In dense video reconstruction, LAL explicitly materializes the attention probabilities so that it can threshold, clip, and renormalize them, introducing additional intermediate tensors. CoANeRV instead expresses temperature modulation as a pre-softmax scale in standard scaled dot-product attention, which permits the fused PyTorch SDPA implementation. Thus, $\tau$ controls coordinate-token retrieval sharpness, whereas the peak-memory reduction comes from combining fused SDPA with spatially tiled coordinate querying rather than from temperature scaling alone.
We compare this design against a video-adapted ANR baseline in Sec.~\ref{sec:experiment}.

\vspace{-8pt}
\section{Experiments}
\label{sec:experiment}
\vspace{-3pt}

\vspace{-5pt}
\subsection{Experimental Setup}
\vspace{-5pt}

The experiments evaluate whether the proposed token-space formulation can replace weight-space video-specificity while preserving reconstruction fidelity and improving memory and encoding efficiency.
Following the ablation in Table~\ref{tab:encode-improvement}, the default model reconstructs videos in normalized YUV space, and the main reconstruction metrics are computed directly in that same space. Conversion back to RGB is used for visualization rather than for the main PSNR/SSIM evaluation.

\textbf{Datasets.}
We follow the evaluation protocol of FastNeRV~\cite{chen2024fast} and report results on three benchmark video datasets.
Kinetics-400 (K400)~\cite{kay2017kinetics} serves as the primary training corpus, containing roughly 240K videos from 400 action classes.
To keep training cost manageable, we use a fixed subset of 10,000 K400 training videos before minimum-frame filtering. This subset was not constructed using class-aware balancing. At runtime, videos with fewer than $F$ decoded frames are excluded, and the remaining videos are shuffled to form training minibatches. Evaluation uses fixed held-out subsets of K400, Something-Something V2~\cite{goyal2017something} (20,000 videos before filtering), and UCF101~\cite{soomro2012UCF101} (3,500 videos). Unless otherwise specified, models are trained on the K400 subset and evaluated on all three held-out sets without dataset-specific fine-tuning. Post-filtering counts are documented in Appendix~\ref{app:dataset_manifest}; exact split membership will accompany the public code release.

\textbf{Implementation details.}
Unless otherwise specified, the dataset loader selects the first $F$ temporally consecutive decoded frames from each video, where $F\in\{4,8,16\}$.
Each selected frame is then resized so that its shorter side is 256 pixels while preserving aspect ratio, followed by a $256\times256$ center crop.
Our convolutional tokenizer and six-layer token former operate at an intermediate width of $D=768$, with 6 attention heads, a per-head dimension of 64, and a feed-forward dimension of 3072; a shared projection then produces $N$ video tokens of dimension $d=72$ for the coordinate decoder.
Training is performed on 1 NVIDIA A800 GPU using AdamW with an initial learning rate of $10^{-4}$ and PyTorch's default weight decay of $10^{-2}$. For the 150-epoch runs, the learning rate remains $10^{-4}$ through epoch 135 and is reduced by a factor of 10 for epochs 136--150; no warmup or cosine schedule is used.
For each minibatch, PSNR is computed from the MSE averaged over all video, frame, channel, and spatial elements. Epoch-level PSNR is the video-count-weighted mean of minibatch values on each device, followed by an average across devices; SSIM uses the same aggregation. Unless a table explicitly states otherwise, the main PSNR and SSIM~\cite{wang2004image}, as well as the appendix MS-SSIM~\cite{wang2003multiscale} and LPIPS~\cite{zhang2018unreasonable}, are evaluated directly on the normalized YUV reconstruction tensors. Appendix~\ref{app:colorspace} gives the exact tensor-level metric protocol.
Unless noted otherwise in the ablations, the reported results correspond to the complete model configuration used in Sec.~\ref{sec:method}.

\textbf{Video-adapted ANR protocol.}
ANR~\cite{zhang2024attention} is originally an image INR method with LAL over spatial queries $(x,y)$.
We adapt it to video as \textit{ANR-V} by extending the queries to $(x,y,t)$ and applying LAL to video-token reconstruction.
In Table~\ref{tab:anr},
ANR-V uses the same video-token interface, coordinate queries, and preprocessing as CoANeRV, but replaces temperature-modulated cross-attention with LAL.
The quality comparison mainly reflects the effect of the coordinate-token retrieval rule, whereas the memory comparison reflects the full decoding implementation, including the use of block-wise query processing in CoANeRV.

\begin{table*}[t!]
\centering
\vspace{-10pt}
\caption{\textbf{CoANeRV vs. feed-forward INR baselines.}
`F' denotes the number of frames.
``INR size'' reports the size of the reconstruction network, while \(\#\hat\theta'\) reports the video-specific representation size: generated video-specific weights for prior methods and video tokens for CoANeRV.
For each frame setting, we report two CoANeRV rows: a 20-epoch checkpoint to highlight convergence speed, and the full 150-epoch result for final accuracy.
Training time is measured in GPU hours.
}
\resizebox{.99\textwidth}{!}{%
\begin{tabular}{@{}lc|ccc|cc|cccc|cccc}
\toprule
\multirow{2}{*}{Methods} & \multirow{2}{*}{F} & \multirow{2}{*}{\makecell{Encoder \\ size}}  & \multirow{2}{*}{\makecell{ INR \\ size $\downarrow$}}  & \multirow{2}{*}{$\#\hat\theta'$ $\downarrow$} & \multirow{2}{*}{Epoch} & \multirow{2}{*}{\makecell{GPU \\ hrs $\downarrow$}} & \multicolumn{4}{c}{PSNR $\uparrow$} & \multicolumn{4}{c}{SSIM $\uparrow$} \\
 &  &  &  &  &  &  & Train & K400 & SthV2 & UCF101 & Train & K400 & SthV2 & UCF101 \\
\midrule
TransINR~\cite{chen2022transinr} & 4 & 48.0M & 99k & 25k & 150 & 63 & 23.7 & 22.1 & 24.6 & 22.1 & 0.659 & 0.631 & 0.728 & 0.622 \\
GINR~\cite{kim2022generalizable} & 4 & 47.6M & 139.4k & 25.6k & 150 & 65 & 24.5 & 23.2 & 25.9 & 23.1 & 0.685 & 0.66 & 0.744 & 0.66 \\
FastNeRV~\cite{chen2024fast} & 4 & 47.6M & 85.6k & 24.1k & {150} & {9} & {26.6} & {26.6} & {29.4} & {26.1} & {0.756} & {0.754} & {0.816} & {0.752} \\
CoANeRV(ours) & 4 & 45.5M & 86.4k & 27k & {20} & {5} & \underline{31.4} & \underline{31.5} & \underline{31.9} & \underline{31.5} & \underline{0.924} & \underline{0.922} & \underline{0.925} & \underline{0.923} \\
CoANeRV(ours) & 4 & 45.5M & 86.4k & 27k & {150} & {39} & \textbf{35.3} & \textbf{34.8} & \textbf{36.0} & \textbf{35.8} & \textbf{0.955} & \textbf{0.951} & \textbf{0.956} & \textbf{0.955} \\
\midrule
TransINR~\cite{chen2022transinr} & 8 & 48.0M & 99k & 25k & 150 & 119 & 22.3 & 20.3 & 22.8 & 20.7 & 0.626 & 0.595 & 0.703 & 0.591 \\
GINR~\cite{kim2022generalizable} & 8 & 47.6M & 139.4k & 25.6k & 150 & 123 & 23.9 & 22.8 & 25.3 & 22.7 & 0.671 & 0.65 & 0.737 & 0.651 \\
FastNeRV~\cite{chen2024fast} & 8 & 47.6M & 85.6k & 24.1k & {150} & {11} & {25.8} & {25.8} & {28.5} & {25.2 }& {0.732} & {0.727} & {0.795} & {0.723 }\\
CoANeRV(ours) & 8 & 46.3M & 86.4k & 27k & {20} & {9} & \underline{28.8} & \underline{28.6} & \underline{28.9} & \underline{28.7} & \underline{0.896} & \underline{0.893} & \underline{0.897} & \underline{0.895} \\
CoANeRV(ours) & 8 & 46.3M & 86.4k & 27k & {150} & {70} & \textbf{33.1} & \textbf{32.1} & \textbf{32.8} & \textbf{33.3} & \textbf{0.939} & \textbf{0.927} & \textbf{0.935} & \textbf{0.938} \\
\midrule
TransINR~\cite{chen2022transinr} & 16 & 48.0M & 99k & 25k & 150 & 234 & 21.5 & 18.4 & 21.1 & 19.2 & 0.615 & 0.555 & 0.678 & 0.561 \\
GINR~\cite{kim2022generalizable} & 16 & 47.6M & 139.4k & 25.6k & 150 & 242 & 22.9 & 21.7 & 24.2 & 21.7 & 0.647 & 0.624 & 0.72 & 0.625 \\
FastNeRV~\cite{chen2024fast} & 16 & 47.6M & 85.6k & 24.1k & {150} & {15} & {23.6} & {23.2} & {25.9} & {22.9} & {0.657} & {0.642} & {0.731} & {0.642} \\
CoANeRV(ours) & 16 & 47.9M & 86.4k & 27k & 20 & {17} & \underline{27.3} & \underline{27.1} & \underline{27.4} & \underline{27.3} & \underline{0.872} & \underline{0.869} & \underline{0.873} & \underline{0.872} \\
CoANeRV(ours) & 16 & 47.9M & 86.4k & 27k & {150} & {128} & \textbf{31.5} & \textbf{30.9} & \textbf{31.1} & \textbf{31.3} & \textbf{0.923} & \textbf{0.917} & \textbf{0.918} & \textbf{0.919} \\
\bottomrule
\end{tabular}
}
\label{tab:compare}
\vspace{-8pt}
\end{table*}

\begin{figure*}[t!]
    \vspace{-3pt}
    \centering
    \includegraphics[width=.9\linewidth]{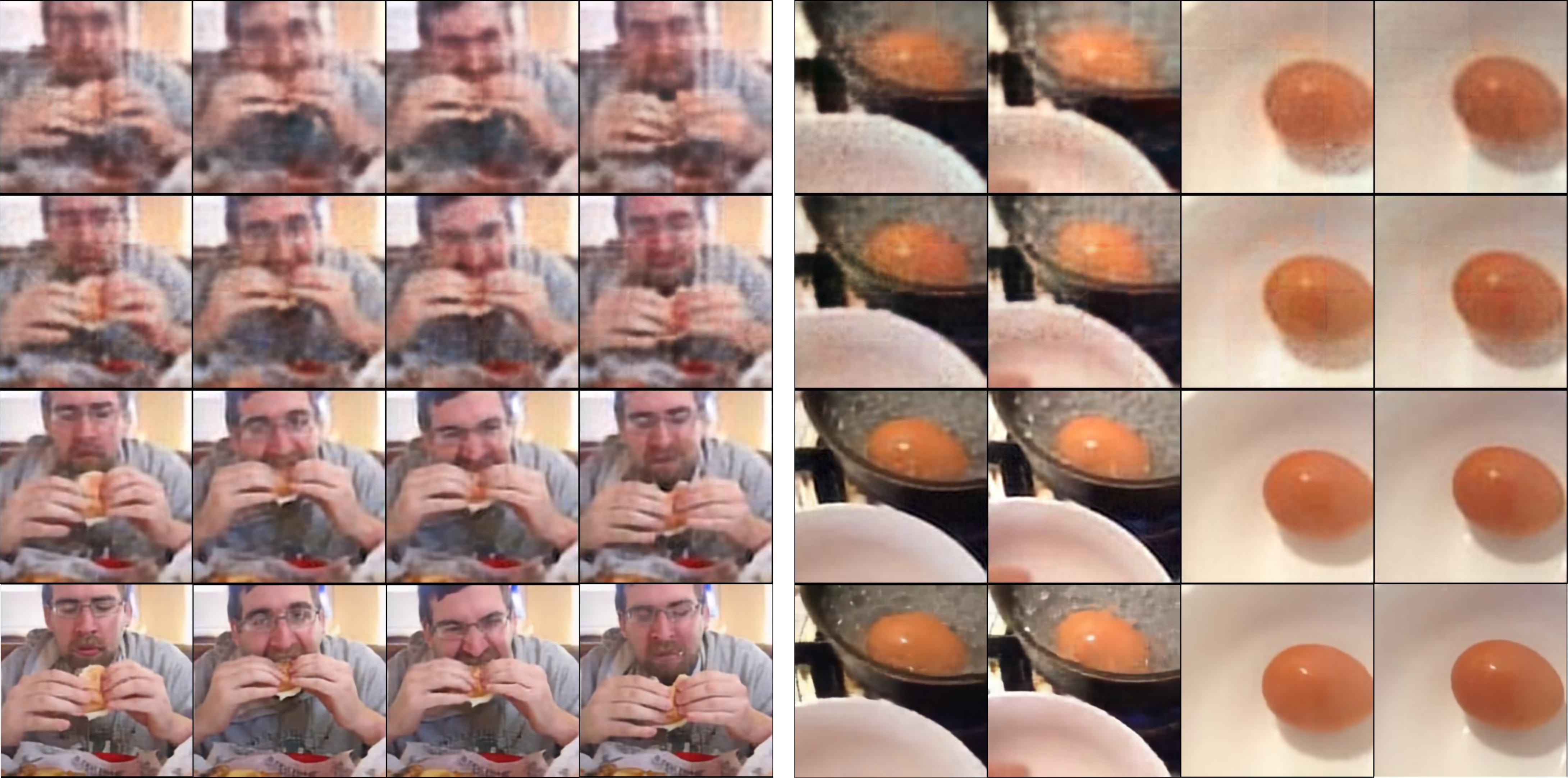}
    \caption{\textbf{Qualitative comparison.}
    Visualizations for feed-forward INR methods: TransINR~\cite{chen2022transinr} (top row), GINR~\cite{kim2022generalizable} (second row), FastNeRV~\cite{chen2024fast} (third row), and CoANeRV (bottom row, ours).
    CoANeRV reconstructs finer details and cleaner structures.
    Best viewed digitally and zoomed in.}
    \label{fig:encode-visualization}
\vspace{-8pt}
\end{figure*}

\vspace{-5pt}
\subsection{Comparison with Feed-Forward INR Baselines}
\vspace{-5pt}

Table~\ref{tab:compare} compares CoANeRV with representative feed-forward NeRV and INR baselines, including TransINR~\cite{chen2022transinr}, GINR~\cite{kim2022generalizable}, and FastNeRV~\cite{chen2024fast}, under the same 4/8/16-frame evaluation protocol.
These results suggest that the improvement is not explained by model size alone, but is closely related to the token-based representation and coordinate-aware decoding design.

\textbf{Reconstruction quality.}
At 4 frames, the full CoANeRV model reaches 35.3 dB on the training split and 34.8/36.0/35.8 dB on K400, SthV2, and UCF101, respectively.
Compared with the strongest feed-forward baseline FastNeRV, this corresponds to gains of 8.7 dB on Train, 8.2 dB on K400, 6.6 dB on SthV2, and 8.7 dB on UCF101.
The same pattern persists at 8 and 16 frames: CoANeRV consistently delivers the best PSNR and SSIM across all three datasets, indicating that the shared coordinate-aware decoder remains effective as the temporal length increases.

\textbf{Convergence behavior.}
The 20-epoch CoANeRV checkpoints are included to separate convergence speed from final accuracy.
Even with only 20 epochs, CoANeRV already surpasses the 150-epoch feed-forward baselines in all frame settings.
For example, at 4 frames the 20-epoch model reaches 31.5 dB on K400, compared with 26.6 dB for FastNeRV; at 16 frames it reaches 27.1 dB on K400, compared with 23.2 dB for FastNeRV.
Training to 150 epochs further improves performance, showing that the method benefits from both faster convergence and a stronger final optimum. We also isolate the representation capacity of the coordinate-conditioned decoder in Appendix~\ref{app:standalone_token_optimization}.

\textbf{Qualitative comparison.}
Figure~\ref{fig:encode-visualization} complements the quantitative results.
Relative to TransINR, GINR, and FastNeRV, CoANeRV reconstructs sharper boundaries, preserves small textures more faithfully, and reduces the oversmoothing artifacts that become prominent in motion-heavy or fine-detail regions.

\begin{table*}[t!]
\centering
\vspace{-5pt}
\caption{\textbf{Comparison with video-adapted ANR.}
ANR-V extends ANR~\cite{zhang2024attention} from spatial queries $(x,y)$ to video queries $(x,y,t)$ using localized attention layers.
CoANeRV uses temperature-modulated cross-attention with block-wise coordinate querying.
Both methods are compared under matched token lengths and video-specific representation sizes.
Memory is the recorded peak GPU footprint for one minibatch and is reported in GB, using $1\,\mathrm{GB}=1024\,\mathrm{MiB}$ throughout the paper.
}
\resizebox{.98\textwidth}{!}{%
\begin{tabular}{@{}lc|cc|ccc|cccc|cccc}
\toprule
\multirow{2}{*}{Methods} & \multirow{2}{*}{F} & \multirow{2}{*}{\makecell{Token \\ length $N$}}   & \multirow{2}{*}{$\#\hat\theta'$ } & \multirow{2}{*}{Epoch} & \multirow{2}{*}{\makecell{Peak memory \\ / batch (GB) $\downarrow$}} & \multirow{2}{*}{\makecell{GPU \\ hrs $\downarrow$}} & \multicolumn{4}{c}{PSNR $\uparrow$} & \multicolumn{4}{c}{SSIM $\uparrow$} \\
 &  &  &  &  &  &  & Train & K400 & SthV2 & UCF101 & Train & K400 & SthV2 & UCF101 \\
\midrule
ANR-V-S & 4 & 384  & 27k & {150} & 18 & {75} & {31.9} & {29.9} & {32.0} & {31.0} & {0.933} & {0.919} & {0.934} & {0.929} \\
CoANeRV-S & 4 & 384  & 27k & {150} & 2.6 & {39} & \textbf{35.3} & \textbf{34.8} & \textbf{36.0} & \textbf{35.8} & \textbf{0.955} & \textbf{0.951} & \textbf{0.956} & \textbf{0.955} \\
ANR-V-M & 4 & 512  & 36k & {150} & 23 & {85} & {32.2} & {30.8} & {33.1} & {32.2} & {0.942} & {0.930} & {0.942} & {0.939} \\
CoANeRV-M & 4 & 512  & 36k & {150} & 2.7 & {41} & \textbf{35.8} & \textbf{35.2} & \textbf{35.6} & \textbf{35.4} & \textbf{0.960} & \textbf{0.952} & \textbf{0.962} & \textbf{0.958} \\
ANR-V-L & 4 & 768  & 54k & {150} & 34 & {132} & {33.6} & {31.2} & {33.4} & {32.9} & {0.947} & {0.934} & {0.946} & {0.946} \\
CoANeRV-L & 4 & 768  & 54k & {150} & 2.7 & {52} & \textbf{37.7} & \textbf{37.1} & \textbf{38.1} & \textbf{38.0} & \textbf{0.970} & \textbf{0.965} & \textbf{0.972} & \textbf{0.973} \\
\midrule
ANR-V-S & 8 & 384  & 27k & {150} & 35 & {178} & {30.9} & {28.3} & {29.8} & {29.3} & {0.921} & {0.901} & {0.914} & {0.907} \\
CoANeRV-S & 8 & 384  & 27k & {150} & 3.9 & {70} & \textbf{33.1} & \textbf{32.1} & \textbf{32.8} & \textbf{33.3} & \textbf{0.939} & \textbf{0.927} & \textbf{0.935} & \textbf{0.938} \\
ANR-V-M & 8 & 512  & 36k & {150} & 46 & {205} & {31.4} & {28.4} & {29.9} & {29.2} & {0.924} & {0.899} & {0.913} & {0.908} \\
CoANeRV-M & 8 & 512  & 36k & {150} & 4.0 & {84} & \textbf{33.6} & \textbf{33.2} & \textbf{33.9} & \textbf{33.6} & \textbf{0.943} & \textbf{0.931} & \textbf{0.939} & \textbf{0.943} \\
ANR-V-L & 8 & 768  & 54k & {150} & 74 & {344} & {32.5} & {32.0} & {32.4} & {32.3} & {0.935} & {0.931} & {0.936} & {0.933} \\
CoANeRV-L & 8 & 768  & 54k & {150} & 4.0 & {92} & \textbf{35.9} & \textbf{34.9} & \textbf{35.9} & \textbf{35.6} & \textbf{0.964} & \textbf{0.960} & \textbf{0.962} & \textbf{0.966} \\
\bottomrule
\end{tabular}
}
\label{tab:anr}
\vspace{-8pt}
\end{table*}

\vspace{-5pt}
\subsection{Memory Efficiency Analysis}
\vspace{-5pt}

Table~\ref{tab:anr} compares CoANeRV with ANR-V, a video-adapted version of ANR~\cite{zhang2024attention}.
Both methods use video tokens and spatio-temporal coordinate queries, but differ in the decoding rule used for coordinate-token retrieval.
ANR-V uses threshold-based localized attention, whereas CoANeRV uses temperature-modulated cross-attention together with block-wise query processing.

The quality comparison mainly reflects the effect of the coordinate-token retrieval rule, while the memory comparison reflects the full dense-decoding implementation, including block-wise query processing in CoANeRV.
Thus, Table~\ref{tab:anr} should be interpreted as a practical decoder-level comparison rather than a fully isolated attention-only ablation.

\textbf{Memory usage.}
CoANeRV consistently operates in the 2.6--4.0 GB range, whereas ANR-V requires 18--74 GB across the same settings.
This corresponds to an 85--95\% reduction in peak per-batch memory.
The gap widens as token length and frame count increase: for example, at 8 frames with the large model, ANR-V-L uses 74 GB while CoANeRV-L remains at 4 GB.

\textbf{Training cost.}
The lower memory footprint is accompanied by substantially reduced training time.
At 4 frames, CoANeRV-L requires 52 GPU hours versus 132 for ANR-V-L while also improving PSNR from 33.6 to 37.7 dB on the training split.
At 8 frames, the same comparison is 92 GPU hours versus 344, again with higher reconstruction quality.
These results show that the proposed decoder is not only more memory-efficient, but also easier to scale in practice.

\textbf{Quality under matched representation size.}
Importantly, the gains are observed under matched token lengths and comparable video-specific representation sizes.
Across all variants, CoANeRV achieves the best PSNR/SSIM while staying within a much smaller memory envelope.
This makes the comparison favorable not only in efficiency, but also in quality at a controlled capacity budget.

\textbf{Capacity--memory--training Pareto.}
For the 8-frame small/medium/large settings, increasing $N=384/512/768$ improves K400 PSNR/SSIM to 32.1/0.927, 33.2/0.931, and 34.9/0.960 at 3.9/4.0/4.0 GB and 70/84/92 GPU hours (Table~\ref{tab:anr}). Memory headroom can therefore be spent on capacity, batching, resolution, or concurrency; Appendix~\ref{app:block_query} gives the exact tiled-decoding formulation.

\textbf{Decoder limitation.}
At 1080p on A100, CoANeRV extracts tokens at 247.0 fps but reconstructs at 7.7 fps, versus 112.8 fps decoding for DCVC-RT. We claim a representation-level quality/storage/encoding/memory trade-off, not the fastest decoder or a complete real-time codec.

\begin{table*}[t!]
\centering
\vspace{-5pt}
\caption{
\textbf{Ablation study.}
Progressive analysis from an ANR-V/LAL attention decoder baseline. The temperature-attention column denotes temperature-scaled standard attention executed through fused SDPA; the reported peak memory reflects the full decoding path, including spatial block querying where enabled.
All memory entries are the recorded peak GPU footprint for one minibatch and are reported in GB, using $1\,\mathrm{GB}=1024\,\mathrm{MiB}$.
}
\resizebox{.98\linewidth}{!}{
\begin{tabular}{@{}ccccc|cc|cccc|cccc@{}}
\toprule
\multirow{2}{*}{\makecell{RGB to\\ YUV}}  & \multirow{2}{*}{\makecell{Block \\ Query}} & \multirow{2}{*}{\makecell{Convolution \\ Tokenizer}} & \multirow{2}{*}{\makecell{Axis-Adaptive- \\ Embedding}} & \multirow{2}{*}{\makecell{Temperature-scaled \\ fused SDPA}} & \multirow{2}{*}{\makecell{Peak memory \\ / batch (GB) $\downarrow$}} & \multirow{2}{*}{\makecell{GPU \\ hrs $\downarrow$}} &  \multicolumn{4}{c|}{PSNR $\uparrow$} & \multicolumn{4}{c}{SSIM $\uparrow$} \\
& & && & & & Train & K400 & SthV2 & UCF101 & Train & K400 & SthV2 & UCF101 \\
 \midrule
& & && & 35&178  & {26.9} & {25.3} & {26.3} & {26.2} & {0.668} & {0.579} & {0.612} & {0.602} \\
\midrule
\checkmark& & &&& 35&178  & {27.8} & {26.5} & {27.4} & {27.2} & {0.726} & {0.693} & {0.708} & {0.701} \\
\checkmark& \checkmark& &&& 18&178  & {27.8} & {26.5} & {27.4} & {27.2} & {0.726} & {0.693} & {0.708} & {0.701} \\
\checkmark & \checkmark& \checkmark & & & 18&172& {28.2} & {27.7} & {28.4} & {28.5} & {0.821} & {0.793} & {0.813} & {0.802} \\
\checkmark& \checkmark && \checkmark &  &18&180& 28.8 & 27.8 & 28.5 & 28.8 & 0.825 & 0.804 & 0.816 & 0.812 \\
\checkmark& \checkmark && & \checkmark &3.9&70& 31.1 & 29.3 & 30.9 & 30.4 & 0.888 & 0.879 & 0.860 & 0.871 \\
\midrule
\checkmark& \checkmark &\checkmark & \checkmark & \checkmark  &3.9&70& \textbf{33.1} & \textbf{32.1} & \textbf{32.8} & \textbf{33.3} & \textbf{0.939} & \textbf{0.927} & \textbf{0.935} & \textbf{0.938} \\
\bottomrule
\end{tabular}
}
\label{tab:encode-improvement}
\vspace{-5pt}
\end{table*}

\vspace{-5pt}
\subsection{Ablation Study}
\vspace{-5pt}

Table~\ref{tab:encode-improvement} studies five design choices in CoANeRV: RGB-to-YUV conversion, block query processing, the convolutional tokenizer, axis-adaptive positional encoding, and temperature-modulated attention.
The first row is an ANR-V/LAL attention decoder baseline, which does not use the proposed auxiliary components.
We then progressively add components, so the table should be interpreted as a progressive design study rather than a fully factorial attribution analysis.

\textbf{Early design choices.}
Starting from the LAL attention decoder baseline (26.9 dB PSNR, 35 GB memory, 178 GPU hours), RGB$\rightarrow$YUV conversion improves reconstruction quality, whereas block query processing preserves the same reconstruction quality while reducing memory from 35 GB to 18 GB.
This shows that the color-space change primarily improves fidelity, whereas query blocking primarily improves memory efficiency.

\textbf{Incremental architectural gains.}
Adding the convolutional tokenizer on top of RGB$\rightarrow$YUV and block query processing further increases K400 PSNR from 26.5 to 27.7 and reduces training time from 178 to 172 GPU hours. Notably, the test set improvements (+1.0–1.3 dB) substantially exceed the training gain, suggesting that the tokenizer may improve transfer to held-out videos.
Replacing this branch with the Axis-Adaptive embedding variant instead yields 27.8 dB on K400 and 28.8 dB on UCF101, showing that explicit spatiotemporal structure also contributes to the final quality.
Replacing the LAL-based decoder with temperature-modulated cross-attention gives the strongest intermediate improvement, raising Train PSNR to 31.1 dB, consistent with Table~\ref{tab:encode-improvement}.
The associated memory reduction should not be attributed to the temperature coefficient itself: it results from expressing temperature as the scale of standard attention, enabling fused SDPA, together with spatial block querying and the removal of LAL's explicit probability thresholding and renormalization.

\textbf{Full model.}
Combining all components produces the final model with 33.1/32.1/32.8/33.3 dB PSNR and 0.939/0.927/0.935/0.938 SSIM on Train/K400/SthV2/UCF101.
Relative to the baseline, this is a cumulative gain of 6.2 dB on the training split together with a nearly 9$\times$ reduction in memory (35 GB $\rightarrow$ 3.9 GB) and a substantial drop in training cost (178 $\rightarrow$ 70 GPU hours).
The ablation therefore supports the view that the final performance emerges from the interaction of multiple design choices rather than from a single isolated component.

\begin{figure}[t!]
    \centering
    \includegraphics[width=0.99\textwidth]{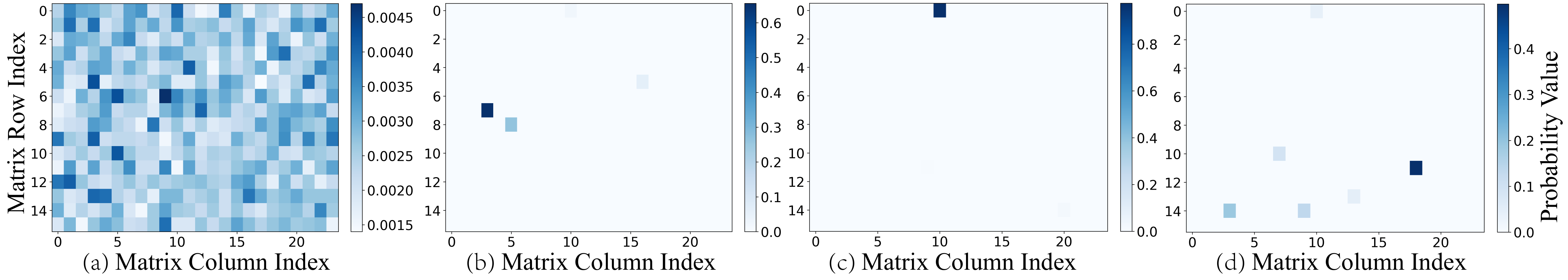}
    \vspace{-5pt}
    \caption{\textbf{Attention pattern analysis.}
    Token probability distributions for
    (a) pre-training and (b) post-training at the same coordinate,
    and (c)--(d) post-training at different coordinates from the same video.
    The figure illustrates the emergence of sharper and more coordinate-specific attention patterns.}
    \label{fig:prob}
    \vspace{-5pt}
\end{figure}

\textbf{Attention mechanism analysis.}
Figure~\ref{fig:prob} visualizes how cross-attention patterns evolve during training.
The pre-training and post-training comparison shows that initially diffuse token distributions become much more concentrated after optimization.
The coordinate-conditioned views further show that different spatial locations emphasize different subsets of tokens, which is consistent with the intended role of the coordinate-aware decoder.
These visualizations provide qualitative support for the temperature-modulated attention design used throughout the model.

\vspace{-5pt}
\subsection{Compression Comparison}
\vspace{-5pt}

\begin{table*}[t!]
\centering
\vspace{-5pt}
\caption{\textbf{Per-video representation-storage comparison.}
Rate-distortion comparison with traditional codecs and FastNeRV under a simple per-video storage protocol.
For CoANeRV and FastNeRV, size reports the quantized method-specific video payload---video tokens for CoANeRV and generated video-specific parameters for FastNeRV---while excluding shared or video-agnostic weights amortized across a video collection.
VPS means videos per second.
}
\renewcommand{\tabcolsep}{1pt}
\resizebox{.75\linewidth}{!}{%
\begin{tabular}{l|c|ccc|ccccc|ccccc}
\toprule
\multirow{2}{*}{}  & \multirow{2}{*}{\makecell{AV1 \\ CRF 60}} & \multicolumn{3}{c|}{H.264} &  \multicolumn{5}{c|}{FastNeRV} & \multicolumn{5}{c}{CoANeRV (ours)} \\
 &   & CRF 35 & CRF 40 & CRF 45 & 8 bits & 7 bits & 6 bits & 5 bits & 4 bits & 8 bits & 7 bits & 6 bits & 5 bits & 4 bits \\
\midrule
Size(KB) $\downarrow$  & 21.9 & 20.4 & 13.1 & {8.7} & 23.7 & 20.7 & 17.7& 14.7 & 11.6 & 26.9 & 23.6 & 20.2 & 16.9 & 13.5 \\
PSNR $\uparrow$  & 32.4 & {32.8} & 30.0 & 27.3 & 28.4 & 28.3 & 28.1 & 27.5 & 25.6& 33.0 & 32.9 & {32.5} & 31.2 & 27.8 \\
SSIM $\uparrow$   & 0.910 & 0.912 & 0.860 & 0.788 & 0.808 & 0.807 & 0.802 & 0.784 & 0.712  & {0.938} & 0.937 & {0.933} & 0.919 & 0.866 \\
\midrule
VPS $\uparrow$  & 313 & 447 & 460 & 485 & \multicolumn{5}{c|} {{583}} & \multicolumn{5}{c} {125}\\
\bottomrule
\end{tabular}
}
\label{tab:nerv-dec-size-quality-speed}
\vspace{-15pt}
\end{table*}

For CoANeRV, we quantize the video-token payload and optionally serialize the resulting symbols with a post-hoc Huffman code for storage accounting. This storage operation is separate from the feed-forward token-extraction throughput reported elsewhere, which excludes quantization, Huffman bitstream construction, and file I/O.
Table~\ref{tab:nerv-dec-size-quality-speed}
evaluates the rate--distortion behavior of token-space representations under a simple storage protocol.
At comparable per-video storage sizes, CoANeRV achieves higher reconstruction quality than FastNeRV; for example, around 20 KB, 6-bit CoANeRV reaches 32.51 dB / 0.933 SSIM, versus 28.3 dB / 0.807 for 7-bit FastNeRV.
Compared with H.264 and AV1, conventional codecs remain highly competitive and decode faster.
We therefore treat this experiment as an analysis of token-space representation storage rather than a claim of state-of-the-art video compression.
The present evaluation uses post-training quantization and optional Huffman symbol coding; it does not evaluate quantization-aware training (QAT), a learned entropy model, or end-to-end rate control.

\vspace{-11pt}
\section{Conclusion and Limitations}
\vspace{-5pt}

We presented CoANeRV, a coordinate-aware token-space formulation for neural video representation.
Rather than instantiating each video as a separate decoder, CoANeRV stores video-specific content in latent tokens and uses a shared coordinate-conditioned decoder to reconstruct spatiotemporal signals.
This perspective decouples video identity from decoder parameters and provides an interface between learned content and coordinate-based reconstruction.
CoANeRV demonstrates an amortized video-specific realization of token-conditioned neural fields when many videos share one decoder. Limitations are the coordinate-attention decoder's throughput and the focus on short clips. The storage study uses post-training quantization and optional Huffman coding, not quantization-aware training, learned entropy, or end-to-end rate control. Future work includes faster decoding, longer-range tokenization, learned token coding, and broader representation transfer.


\bibliographystyle{plainnat}
\bibliography{example_paper}

\appendix
\clearpage
\section{Appendix}

This appendix provides additional analysis for CoANeRV as a coordinate-aware token-space neural video representation.
We first clarify the tokenizer and positional encoding used to construct video tokens and coordinate queries.
We then analyze how persistent shared decoding changes storage and encoding behavior compared with weight-space representations, and how block-wise querying reduces peak attention memory.
Finally, we provide additional empirical results, including standalone token optimization, perceptual metrics, scalability analysis, and compression-oriented evaluation.

\newcounter{appendixprop}
\renewcommand{\theappendixprop}{\arabic{appendixprop}}

\subsection{Video Tokenization Methods}
\label{app:tokenization}

Given an input video tensor $V \in \mathbb{R}^{T \times C \times H \times W}$, we compare two tokenization strategies: the standard unfold$+$linear projection and our strided convolutional tokenizer. Both partition the video into non-overlapping $P\times P$ spatial patches and produce patch tokens with the intermediate token-former width $D$; the final stored video-token width $d$ is introduced only after token formation and projection.

\paragraph{Baseline unfold tokenizer.}
Omitting the batch dimension for clarity, let \(V \in \mathbb{R}^{T\times C\times H\times W}\).
In implementation, we reshape the video to \((BT)\times C\times H\times W\) before applying 2D patch extraction or strided convolution.
The unfold operator extracts non-overlapping patches:
\begin{equation}
V_{\mathrm{unfold}} = \texttt{unfold}(V') \in \mathbb{R}^{T \times (C P^2) \times L},
\end{equation}
where $L=(H/P)(W/P)$ is the number of patches per frame. After reshaping into a matrix of flattened patches, a learned linear map projects each patch to $D$ dimensions:
\begin{equation}
Z = \mathrm{Linear}(C P^2 \rightarrow D)(V_{\mathrm{flat}}).
\end{equation}

\paragraph{Convolutional tokenizer.}
Our tokenizer instead uses a strided convolution:
\begin{equation}
V_{\mathrm{conv}} = \mathrm{Conv2d}(C \rightarrow D,\; \mathrm{kernel\_size}=P,\; \mathrm{stride}=P)(V'),
\end{equation}
which produces $V_{\mathrm{conv}} \in \mathbb{R}^{T \times D \times H' \times W'}$ with $H'=H/P$ and $W'=W/P$. Flattening the spatial dimensions yields the token sequence
\begin{equation}
Z \in \mathbb{R}^{(T H' W') \times D}.
\end{equation}

\textbf{Equivalence to patchwise linear projection.}
For non-overlapping patches, a convolution with kernel size $P$ and stride $P$ is algebraically equivalent to applying the same linear map to every flattened $P\times P$ patch.
Let $\mathcal{P}_{t,u,v} \in \mathbb{R}^{C\times P\times P}$ denote the patch extracted from frame $t$ at spatial index $(u,v)$. For output channel $r$, the convolution computes
\begin{equation}
Z_{t,u,v,r} = \langle W_r, \mathcal{P}_{t,u,v} \rangle + b_r,
\end{equation}
where $W_r \in \mathbb{R}^{C\times P\times P}$ is the convolution kernel and $b_r$ is the bias. After vectorization,
\begin{equation}
Z_{t,u,v,r} = \langle \mathrm{vec}(W_r), \mathrm{vec}(\mathcal{P}_{t,u,v}) \rangle + b_r,
\end{equation}
which is exactly a linear projection of the flattened patch. Hence the convolutional tokenizer is not a heuristic replacement for unfold; it implements the same patchwise linear family while avoiding explicit patch materialization. The detailed tensor shapes are shown in Table~\ref{tab:conv}.

The gain does not come from changing the class of local patch projections, but from implementing that class in a memory-efficient and hardware-friendly form that is better aligned with modern vision architectures~\cite{dosovitskiy2020image,arnab2021vivit,liu2021swin,lecun1998gradient,krizhevsky2012imagenet}.

\subsection{Relation to Token-Conditioned Neural Fields}
\label{app:related_token_fields}

Table~\ref{tab:token_field_comparison} makes the novelty boundary explicit. Generic per-instance tokens, shared coordinate decoders, and coordinate-to-token cross-attention are established components of token-conditioned neural fields. CoANeRV's contribution is their coupled realization and evaluation for amortized video representation, not any one of these generic components in isolation.

\begin{table*}[t]
\centering
\footnotesize
\caption{Technical positioning relative to closely related token-conditioned neural fields. ``Not evaluated'' describes the scope of the cited paper and does not assert that the method could not be extended.}
\label{tab:token_field_comparison}
{\setlength{\tabcolsep}{3pt}
\renewcommand{\arraystretch}{1.08}
\begin{tabular}{@{}p{0.11\textwidth}|p{0.17\textwidth}|p{0.13\textwidth}|p{0.23\textwidth}|p{0.26\textwidth}@{}}
\toprule
Method & New-instance path & Query domain & Retrieval/decoder & Video-specific distinction \\
\midrule
LAG-INR & Feed-forward Transformer tokens & 2D coordinates; 6D rays & Local aggregation and multi-band modulation & Temporal modeling and exact dense-video tiling not evaluated \\
Tokenized Neural Fields & Gradient-fitted tokens; frozen decoder & 1D/2D coordinates; 6D rays & Standard cross-attention and shared MLP & No amortized encoder or evaluated temporal axis \\
LH-NeF & Hierarchical feed-forward tokenizer & General metric domains & Group routing, Gaussian cross-attention, and FiLM & No evaluated video temporal modeling \\
CoANeRV & Feed-forward video tokenizer and token former & Continuous $(x,y,t)$ & Temperature-scaled retrieval and shared MLP & Axis-adaptive encoding and exact full-token-bank video tiling \\
\bottomrule
\end{tabular}}
\end{table*}

LAG-INR~\cite{lee2023laginr} is the closest architectural precedent because it already combines feed-forward latent tokens, coordinate cross-attention, and a shared INR decoder. Tokenized Neural Fields~\cite{haider2025tokenized} differs in fitting the tokens of an unseen instance by gradient optimization, whereas CoANeRV predicts video tokens in one encoder pass. LH-NeF~\cite{urbano2026hierarchy} is a concurrent June 2026 preprint that develops modality-agnostic hierarchical locality routing; its contribution and memory comparison are technically distinct from dense video-query tiling. Across these comparisons, CoANeRV specifically targets complete-clip token formation, anisotropic spatio-temporal querying, and collection-level video reconstruction, storage, encoding, and memory trade-offs.

\subsection{Axis-Adaptive Positional Encoding}
\label{app:positional_encoding}

A standard Fourier-style encoding allocates the same number of frequencies to each coordinate axis. For video reconstruction, this isotropic design is unnecessarily restrictive because spatial and temporal axes exhibit different spectral behavior: fine texture and edges require richer spatial frequencies, whereas temporal dynamics are typically smoother due to motion coherence and finite frame rates.

Let the baseline budget be $k \in 3\mathbb{Z}$ frequencies per axis. We reallocate the same total budget according to
\begin{equation}
2k_s + k_t = 3k,
\end{equation}
with spatial allocation
\begin{equation}
k_s =  \frac{4k}{3} ,
\end{equation}
and temporal allocation
\begin{equation}
k_t = 3k - 2k_s =  \frac{k}{3} .
\end{equation}
The associated frequency bands are
\begin{equation}
\mathbf{w}_s = \left[\sigma_s^{\frac{i}{k_s-1}}\right]_{i=0}^{k_s-1}, \qquad
\mathbf{w}_t = \left[\sigma_t^{\frac{i}{k_t-1}}\right]_{i=0}^{k_t-1},
\end{equation}
which define the encoding
\begin{equation}
\gamma(x,y,t) = [\sin(\pi x\mathbf{w}_s),\cos(\pi x\mathbf{w}_s),\sin(\pi y\mathbf{w}_s),\cos(\pi y\mathbf{w}_s),\sin(\pi t\mathbf{w}_t),\cos(\pi t\mathbf{w}_t)].
\end{equation}

\textbf{Budget preservation and anisotropic allocation.}
The axis-adaptive encoding preserves the same total embedding dimensionality as the uniform baseline while allocating a larger fraction of frequency components to the spatial axes.

The uniform baseline uses $k$ sine and $k$ cosine components for each of the three axes, resulting in total dimensionality $6k$.
Our encoding uses $2k_s$ trigonometric components for $x$, $2k_s$ for $y$, and $2k_t$ for $t$, for a total of
\begin{equation}
2k_s + 2k_s + 2k_t = 2(2k_s+k_t)=6k.
\end{equation}
Thus the embedding dimension is unchanged. Since $k_s>k_t$, the spatial axes receive more logarithmically spaced frequency components than the temporal axis under the same total budget.

\begin{table}[t]
\centering
\small
\caption{Data shapes in the convolutional video tokenizer.}
\begin{tabular}{lcc}
\toprule
Stage & Shape & Example \\
\midrule
Input video & \(T \times C \times H \times W\) & \(4 \times 3 \times 256 \times 256\) \\
After Conv2D & \(T \times D \times H' \times W'\) & \(4 \times 768 \times 16 \times 16\) \\
Patch tokens & \(N_p \times D\) & \(1024 \times 768\) \\
Video tokens & \(N \times d\) & \(384 \times 72\) \\
\bottomrule
\end{tabular}
\label{tab:conv}
\end{table}

\paragraph{Choice of frequency scales.}
Following the Nyquist intuition for sampled video signals, we set
\begin{equation}
\sigma_{\mathrm{spatial}} = 2\max(H,W), \qquad \sigma_{\mathrm{temporal}} = 2T,
\end{equation}
so that the encoding range scales with the effective bandwidth of the underlying video grid~\cite{tancik2020fourier,mildenhall2021nerf}. In practice, $\sigma_s=512$ provides a stable balance between local detail sensitivity and coarse-scale robustness for $256\times256$ videos.

\paragraph{Interpretation.}
This design does not claim a new universal positional encoding theorem. Rather, it formalizes a modeling prior specific to videos: under a fixed embedding budget, allocating more frequencies to spatial coordinates and fewer to temporal coordinates is consistent with the anisotropic spectral structure of natural videos~\cite{cover2006elements}.

\subsection{Persistent Weights and Encoding Efficiency}
\label{app:persistent}

The central distinction between CoANeRV and prior weight-generation methods is where video-specific information is stored. In per-video optimization or hypernetwork-based decoding, video-specificity lives in \emph{weight space}. In CoANeRV, the decoder parameters remain fixed and video-specificity is carried by compact tokens.

Let $p_{\mathrm{dec}}$ denote the number of decoder parameters and let each video be represented by $N$ tokens of dimension $d$.
Shared encoder parameters are treated as a one-time cost and do not affect the per-video storage scaling.

\textbf{Storage scaling across multiple videos.}
For a collection of $M$ videos, storing separate per-video decoders scales as $\mathcal{O}(Mp_{\mathrm{dec}})$, whereas using a shared persistent decoder with video-specific tokens scales as
\begin{equation}
\mathcal{O}(p_{\mathrm{dec}} + MNd).
\end{equation}
For sufficiently many videos and compact tokens satisfying $Nd \ll p_{\mathrm{dec}}$, the token-space representation has lower per-video storage growth.

A per-video decoder requires one set of decoder weights for each video, hence total storage $Mp$. In CoANeRV, the persistent decoder is stored once, contributing $p$, while each video contributes only its token set of size $Nd$. Therefore the total cost is $p + MNd$. The difference is
\begin{equation}
Mp - (p + MNd) = (M-1)p - MNd,
\end{equation}
which is positive whenever $Nd < \frac{M-1}{M}p$. For moderate and large $M$, this is well approximated by $Nd < p$.

For sufficiently many videos and compact tokens satisfying $Nd \ll p$, token-space storage provides lower per-video storage growth than storing separate decoder weights.

\textbf{Encoding complexity relative to per-video optimization.}
Suppose a gradient-based video representation requires $E$ optimization steps, each costing $\mathcal{C}_{\mathrm{step}}$, while CoANeRV produces tokens by a single forward pass with cost $\mathcal{C}_{\mathrm{enc}}$. Then the per-video encoding costs are
\begin{equation}
\mathcal{O}(E\,\mathcal{C}_{\mathrm{step}}) \quad \text{vs.} \quad \mathcal{O}(\mathcal{C}_{\mathrm{enc}}),
\end{equation}
So, for fixed relative costs $\mathcal{C}_{\mathrm{step}}/\mathcal{C}_{\mathrm{enc}}$, the speedup grows linearly with the number of optimization steps.
The gradient-based method must repeatedly evaluate forward and backward passes for $E$ iterations, leading to cost proportional to $E\,\mathcal{C}_{\mathrm{step}}$. CoANeRV replaces this iterative optimization with one encoder pass followed by token storage, whose cost is independent of $E$. Hence the asymptotic ratio is proportional to $E$.
The speedup ratio is $\mathcal{O}(E\mathcal{C}_{\mathrm{step}}/\mathcal{C}_{\mathrm{enc}})$; hence, when the relative step and encoder costs are fixed, it grows linearly with the number of optimization steps.

\begin{figure}[t!]
    \centering
    \includegraphics[width=0.8\textwidth]{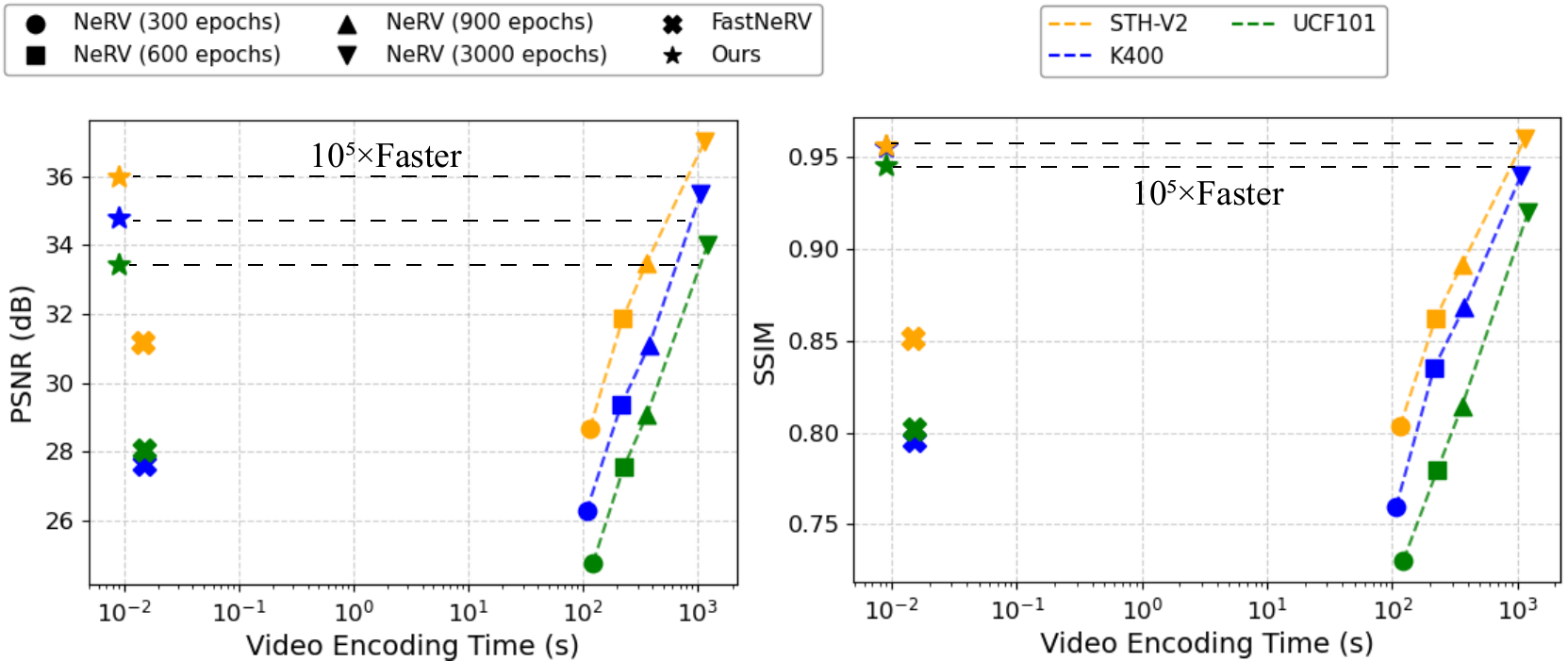}
    \caption{Encoding time comparison between CoANeRV and gradient-based NeRV optimization.}
    \label{fig:encoding}
\end{figure}

This provides a simple explanation for the observed encoding-time gap in Figure~\ref{fig:encoding}.
The observed $10^5\times$ speedup is consistent with this distinction, since feed-forward token prediction removes the per-video optimization loop. The encoding speedup is measured against gradient-based per-video optimization methods. Compared with other feed-forward encoder-based methods, CoANeRV has similar encoding latency; its main advantage in this regime is higher reconstruction quality and improved memory efficiency during coordinate decoding.

\subsection{Standalone Token Optimization Analysis}
\label{app:standalone_token_optimization}

CoANeRV is designed as a feed-forward video representation framework, where video-specific tokens are produced by an encoder in a single forward pass. To further isolate the representation capacity of the coordinate-conditioned decoder, we evaluate an optimization-based variant of CoANeRV. In this variant, all decoder parameters are frozen after training, and only the video-specific tokens are randomly initialized and optimized for each test video. This setting removes the feed-forward encoder while preserving the decoder architecture, providing an architecture-matched comparison between amortized token inference and per-video token optimization.

Specifically, for each test video \(V\), we initialize a token matrix
\(\widetilde{\mathbf{T}}_v \in \mathbb{R}^{N \times d}\) from a Gaussian distribution and optimize it by minimizing
\begin{equation}
    \min_{\widetilde{\mathbf{T}}_v}
    \sum_{q\in\Omega}
    \left\|
    V(q) -
    D_{\psi}\bigl(\gamma(q), \widetilde{\mathbf{T}}_v\bigr)
    \right\|_2^2,
\end{equation}
where \(D_{\psi}\) denotes the frozen CoANeRV decoder,
\(\gamma(\cdot)\) is the coordinate positional encoding, and
\(\Omega\) is the set of queried spatio-temporal coordinates.
This experiment differs from the main feed-forward setting, but it allows us to examine whether the shared decoder itself can serve as a standalone coordinate-based representation network when paired with optimized video tokens.

\begin{table}[t]
\centering
\small
\caption{Standalone token optimization analysis under the 8-frame setting. For CoANeRV, the decoder is frozen and only randomly initialized video tokens are optimized for each test video. Time denotes per-video optimization time.}
\label{tab:standalone_token_optimization}
\begin{tabular}{lcccccccccc}
\toprule
\multirow{2}{*}{Method} & \multirow{2}{*}{Frames} & \multirow{2}{*}{Iter.} & \multirow{2}{*}{Time (s)}
& \multicolumn{3}{c}{PSNR $\uparrow$}
& \multicolumn{3}{c}{SSIM $\uparrow$} \\
\cmidrule(lr){5-7} \cmidrule(lr){8-10}
& & & & K400 & SthV2 & UCF101 & K400 & SthV2 & UCF101 \\
\midrule
NeRV & 8 & 300  & 122  & 24.78 & 24.13 & 24.75 & 0.73 & 0.70 & 0.72 \\
NeRV & 8 & 600  & 230  & 27.55 & 27.25 & 27.11 & 0.81 & 0.80 & 0.79 \\
NeRV & 8 & 900  & 373  & 29.82 & 29.89 & 29.29 & 0.85 & 0.87 & 0.86 \\
NeRV & 8 & 2400 & 942  & 33.64 & 33.73 & 33.17 & 0.90 & 0.91 & 0.89 \\
\midrule
CoANeRV & 8 & 300  & 254  & 24.94 & 24.91 & 24.89 & 0.78 & 0.77 & 0.75 \\
CoANeRV & 8 & 600  & 512  & 29.08 & 28.83 & 28.75 & 0.89 & 0.85 & 0.84 \\
CoANeRV & 8 & 900  & 764  & 32.23 & 32.55 & 32.11 & 0.95 & 0.93 & 0.92 \\
CoANeRV & 8 & 2400 & 2132 & 35.64 & 34.99 & 34.19 & 0.97 & 0.96 & 0.94 \\
\bottomrule
\end{tabular}
\end{table}

Table~\ref{tab:standalone_token_optimization} shows that the frozen CoANeRV decoder remains an effective coordinate-based representation network when paired with optimized video tokens. At matched optimization iterations, CoANeRV generally achieves higher reconstruction quality than NeRV. For example, at 900 iterations on K400, CoANeRV reaches 32.23 dB PSNR, compared with 29.82 dB for NeRV. With sufficient optimization, the standalone CoANeRV variant reaches 34--35 dB PSNR across the evaluated datasets.

This analysis also clarifies the source of the encoding efficiency in the main feed-forward setting. The optimization-based CoANeRV variant requires approximately 764 seconds per video at 900 iterations to reach 32.23/32.55/32.11 dB PSNR on K400/SthV2/UCF101. In contrast, the feed-forward CoANeRV encoder produces video tokens in approximately \(6\) ms per video and achieves comparable reconstruction quality under the same 8-frame setting. Thus, the feed-forward encoder can be interpreted as amortizing the otherwise expensive per-video token optimization process, yielding an encoding acceleration on the order of \(10^5\times\) relative to optimization-based token fitting.

\subsection{Temperature-Modulated Attention}
\label{app:attention}

Let $a_i$ denote the pre-softmax attention logit between a query coordinate and token $i$. Temperature-modulated attention defines
\begin{equation}
p_i(\tau) = \frac{\exp(a_i/\tau)}{\sum_j \exp(a_j/\tau)},
\end{equation}
with attention output
\begin{equation}
\mathrm{Attn}_{\tau}(Q,K,V) = p(\tau)^{\top}V.
\end{equation}
This formulation yields two basic and useful properties.

\textbf{Temperature controls relative selectivity.}
For any two logits $a_i > a_j$,
\begin{equation}
\frac{p_i(\tau)}{p_j(\tau)} = \exp\!\left(\frac{a_i-a_j}{\tau}\right).
\end{equation}
Hence decreasing $\tau$ strictly amplifies the ratio between preferred and non-preferred tokens, making attention more selective.
Since \(a_i-a_j>0\), this ratio increases as \(\tau\) decreases.

\textbf{Attention entropy is monotone in temperature.}
Define the entropy
\begin{equation}
H(\tau) = -\sum_i p_i(\tau)\log p_i(\tau).
\end{equation}
If the logits are not all equal, then
\begin{equation}
\frac{dH}{d\tau} = \frac{\mathrm{Var}_{p(\tau)}[a]}{\tau^3} > 0.
\end{equation}
Therefore increasing $\tau$ makes the attention distribution smoother, while decreasing $\tau$ makes it sharper.

\begin{figure*}[!t]
\centering
\includegraphics[width=0.58\textwidth]{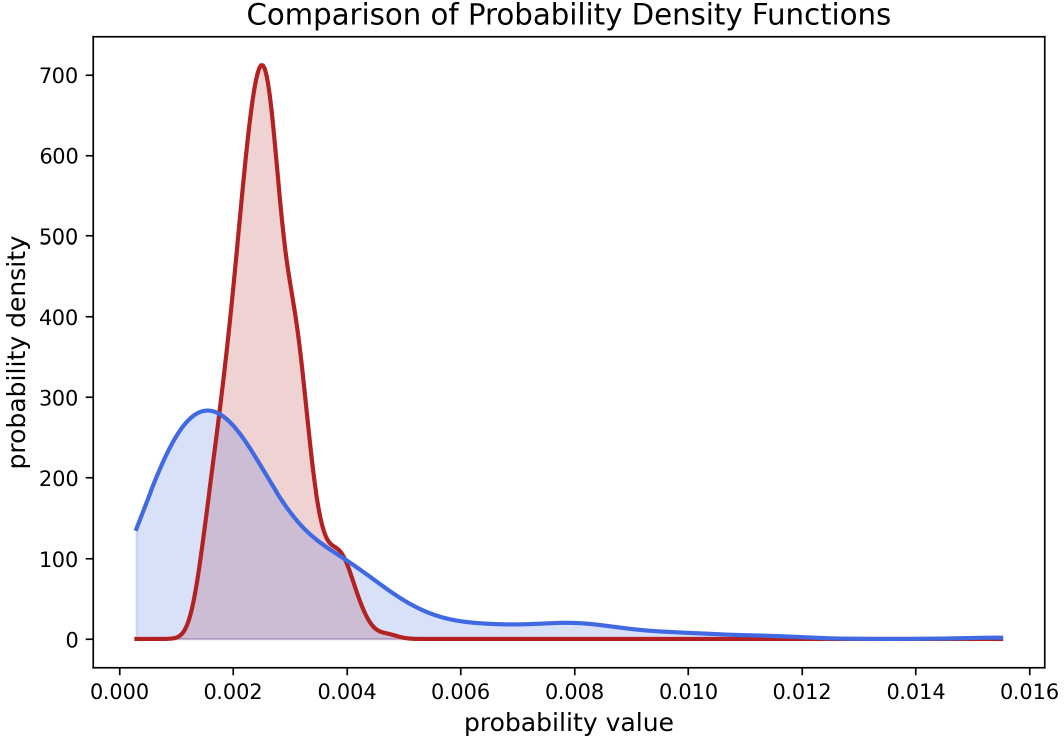}
\caption{\textbf{Attention Weight Distribution with Different Temperature Parameters.} Distribution of attention weights when querying spatiotemporal coordinates through video tokens before training.
Red curve (\(\tau=0.4\)) shows a sharper distribution over tokens, while blue curve (\(\tau=1.0\)) exhibits smoother and more dispersed attention weights.
}
\label{fig:density}
\end{figure*}

Let $\beta=1/\tau$. Then $p_i(\tau)=\exp(\beta a_i)/Z(\beta)$ and
\begin{equation}
H(\beta)=\log Z(\beta)-\beta\,\mathbb{E}_{p(\beta)}[a].
\end{equation}
Differentiating with respect to $\beta$ gives
\begin{equation}
\frac{dH}{d\beta} = -\beta\,\mathrm{Var}_{p(\beta)}[a] \le 0.
\end{equation}
Using $\beta=1/\tau$ and $d\beta/d\tau=-1/\tau^2$ yields
\begin{equation}
\frac{dH}{d\tau} = \frac{\mathrm{Var}_{p(\tau)}[a]}{\tau^3} \ge 0,
\end{equation}
with strict inequality whenever the logits are not all identical. This analysis only characterizes the effect of temperature on attention sharpness for fixed logits; it does not by itself guarantee improved reconstruction.
As shown in Figure~\ref{fig:density}, temperature provides continuous control over attention sharpness. Smaller $\tau$ yields more selective token aggregation, whereas larger $\tau$ produces smoother context aggregation.

\begin{figure}[t!]
    \centering
    \includegraphics[width=0.94\textwidth]{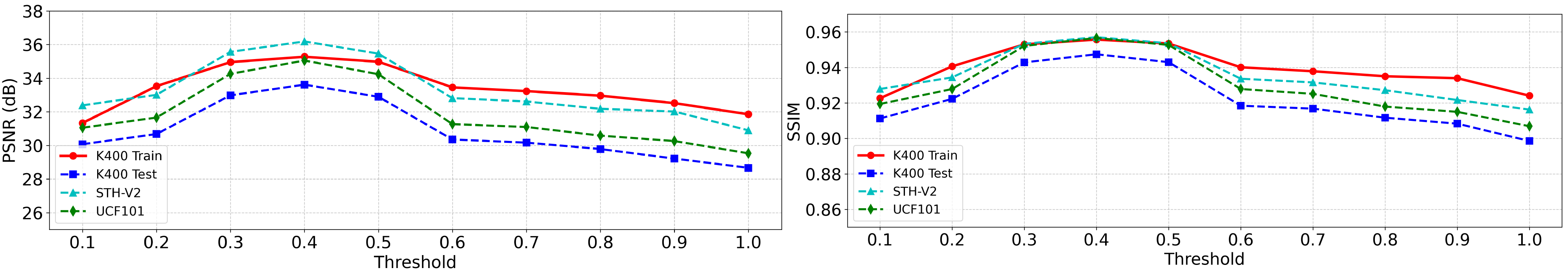}
    \vspace{-5pt}
    \caption{\textbf{Temperature Coefficient.}
PSNR/SSIM results across training and test sets for different $\tau$ values. Both metrics peak at $\tau = 0.4$, indicating optimal attention sharpness. Lower $\tau$ causes over-concentration; higher $\tau$ leads to diffuse attention.}
    \label{fig:threshold}
    \vspace{-10pt}
\end{figure}

\textbf{Temperature Coefficient Analysis.}
We evaluate $\tau \in [0.1,1.0]$ to study the effect of attention sharpness.
As shown in Figure~\ref{fig:threshold}, overly small values such as $\tau=0.1$ produce excessively concentrated attention, while larger values such as $\tau=1.0$ lead to diffuse aggregation.
The best PSNR/SSIM is obtained at $\tau=0.4$, which is used in all main experiments.

\subsection{Block Query Processing and Peak Memory}
\label{app:block_query}

Let $\Omega$ denote the $T\times H\times W$ reconstruction grid, let $N$ be the token length, and let $s$ be the spatial tile side length. Without tiling, a dense attention layer processes $|\Omega|N=THWN$ coordinate-token score pairs.

The implementation splits the height and width axes into non-overlapping $s\times s$ tiles and keeps all $T$ temporal positions inside each tile. For $s\mid H$ and $s\mid W$,
\begin{equation}
\Omega=\bigcup_{b=1}^{B_s}\Omega_b,
\qquad
B_s=\frac{H}{s}\frac{W}{s},
\qquad
|\Omega_b|=Ts^2\equiv M.
\end{equation}
For every tile, its $Ts^2$ coordinate queries attend to the same complete token set $\mathbf{T}_v$; the token memory is not spatially truncated.

\textbf{Exactness of spatially tiled decoding.}
For a fixed query $q$, the output depends only on its own query projection $Q(q)$ and the token-derived keys and values $(K,V)$ computed from $\mathbf{T}_v$:
\begin{equation}
\hat{V}(q) = \mathrm{MLP}\!\left(\mathrm{Softmax}\!\left(\frac{Q(q)K^{\top}}{\tau\sqrt{d_h}}\right)V\right).
\end{equation}
No operation couples different coordinate queries. Therefore, evaluating $q$ alone, inside an $s\times s$ spatial tile containing all $T$ frames, or inside the full query grid yields the same mathematical output. Fused and mixed-precision kernels may introduce negligible floating-point variation, but tiling does not change the decoder function.

\textbf{Peak-memory reduction.}
At most one tile containing $Ts^2$ queries is processed at a time, so the peak attention-memory order becomes $\mathcal{O}(Ts^2N)$. Summing over all $B_s=(H/s)(W/s)$ spatial tiles gives
\begin{equation}
B_s\,Ts^2N=THWN=|\Omega|N,
\end{equation}
so the total arithmetic order remains $\mathcal{O}(|\Omega|N)$. In the reported configuration, $s=64$; consequently, the corresponding query count is $M=T\cdot64^2$, not $M=64$.

\subsection{Color Space Optimization}
\label{app:colorspace}

\begin{figure*}[!t]
\centering
\includegraphics[width=0.98\textwidth]{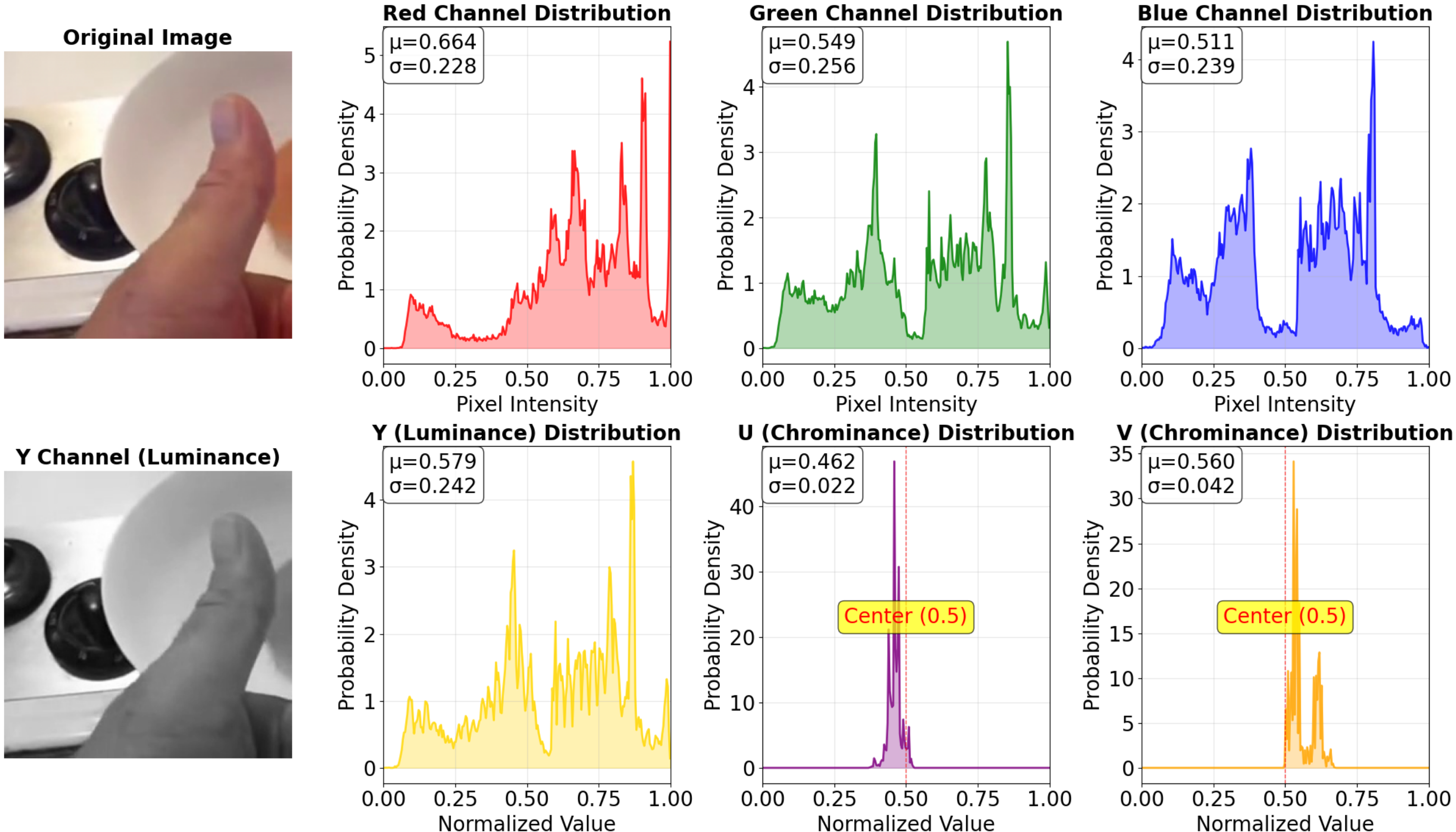}
\caption{\textbf{RGB vs. YUV distributions.} RGB channels occupy the full normalized range, whereas the YUV chrominance channels are strongly concentrated around $0.5$.}
\label{fig:YUV}
\end{figure*}

We adopt the YUV color space instead of RGB to exploit the statistical asymmetry of luminance and chrominance. The input pipeline uses the following full-range BT.601-style transform~\cite{bt2011studio}, with normalized RGB inputs and chrominance centered at $0.5$:
\begin{equation}
\begin{split}
Y &= 0.299R + 0.587G + 0.114B,\\
U &= -0.169R - 0.331G + 0.500B + 0.5,\\
V &= 0.500R - 0.419G - 0.081B + 0.5.
\end{split}
\end{equation}
No clipping or chroma subsampling is applied before the three channels are provided to the model.
The luminance channel $Y$ retains most structural information, whereas $U$ and $V$ are concentrated around the neutral value $0.5$, as shown in Figure~\ref{fig:YUV}.
This suggests that YUV provides a reconstruction space whose channel statistics are better aligned with natural video structure.

The model predicts all three full-resolution YUV channels jointly, and the main reconstruction metrics are computed before RGB conversion. Within each minibatch, MSE is averaged over the complete $B\times F\times3\times H\times W$ tensor and converted to PSNR. SSIM is evaluated over the same batched five-dimensional tensor and averaged by the metric implementation. MS-SSIM and LPIPS fold the video and frame axes into a $BF\times3\times H\times W$ frame batch and remain in normalized YUV space. Epoch-level metrics are video-count-weighted means of minibatch values on each device, followed by an average across devices. Rows without RGB-to-YUV conversion in Table~\ref{tab:encode-improvement} use the same aggregation in RGB space. YUV-to-RGB conversion is reserved for visualizations.
The YUV conversion empirically improves reconstruction, as shown in Table~\ref{tab:encode-improvement}.
This is consistent with both the statistics of natural video and perceptual sensitivity to luminance structure~\cite{jahne2005digital,poynton2012digital,salomon2002data,wandell1995foundations}.

\subsection{Perceptual Reconstruction Quality}
\label{app:perceptual_quality}

PSNR and SSIM are widely used for evaluating neural video reconstruction, but they mainly measure pixel-level fidelity and may not fully reflect perceptual quality. To further verify whether the reconstruction gains of CoANeRV correspond to visually meaningful improvements, we additionally evaluate MS-SSIM and LPIPS~(AlexNet) on the test sets. MS-SSIM measures multi-scale structural similarity, while LPIPS evaluates perceptual feature-space distance, where lower values indicate better perceptual quality.

\begin{table}[t]
\centering
\small
\caption{Perceptual reconstruction quality under the 4-frame setting. CoANeRV consistently improves MS-SSIM and reduces LPIPS across all evaluated datasets.}
\label{tab:perceptual_quality}
\begin{tabular}{lccccccc}
\toprule
\multirow{2}{*}{Method} & \multirow{2}{*}{Frames}
& \multicolumn{3}{c}{MS-SSIM $\uparrow$}
& \multicolumn{3}{c}{LPIPS $\downarrow$} \\
\cmidrule(lr){3-5} \cmidrule(lr){6-8}
& & K400 & SthV2 & UCF101 & K400 & SthV2 & UCF101 \\
\midrule
TransINR & 4 & 0.787 & 0.868 & 0.843 & 0.565 & 0.361 & 0.411 \\
GINR     & 4 & 0.829 & 0.895 & 0.876 & 0.494 & 0.304 & 0.331 \\
FastNeRV & 4 & 0.921 & 0.954 & 0.941 & 0.267 & 0.139 & 0.167 \\
\textbf{CoANeRV} & 4 & \textbf{0.981} & \textbf{0.983} & \textbf{0.984}
& \textbf{0.052} & \textbf{0.044} & \textbf{0.034} \\
\bottomrule
\end{tabular}
\end{table}

As shown in Table~\ref{tab:perceptual_quality}, CoANeRV consistently achieves the best perceptual reconstruction quality across all three datasets. Compared with the strongest baseline FastNeRV, CoANeRV improves MS-SSIM from 0.921--0.954 to 0.981--0.984 and reduces LPIPS from 0.139--0.267 to 0.034--0.052. These results indicate that the reconstruction improvements are not limited to pixel-wise distortion metrics, but also correspond to better structural and perceptual fidelity. The qualitative comparisons further support this observation, where CoANeRV preserves sharper object boundaries, more coherent textures, and more accurate colors.

\subsection{Computational Complexity Analysis}
\label{app:complexity}

Table~\ref{tab:complexity_comparison} reports the measured FLOPs and MACs for reconstructing a single frame. The table should be read together with the theoretical discussion above: CoANeRV does not attempt to minimize raw per-frame arithmetic at all costs, but rather to improve the quality--memory--encoding trade-off relative to existing feed-forward and transformer-based baselines.

\begin{table}[ht]
\centering
\caption{Computational complexity comparison for single-frame reconstruction. FLOPs and MACs are measured in billions (G).}
\label{tab:complexity_comparison}
\begin{tabular}{lccc}
\toprule
\textbf{Method} & \textbf{FLOPs (G)} & \textbf{MACs (G)} & \textbf{Architecture Type} \\
\midrule
NeRV & 1.30 & 0.65 & Convolutional \\
ANR-V & 239.73 & 119.45 & LAL-based attention \\
CoANeRV (Ours) & 31.92 & 15.95 & Temperature-modulated attention \\
\bottomrule
\end{tabular}
\end{table}

\paragraph{Interpretation.}
NeRV is the cheapest per frame because it uses a direct convolutional decoder with no attention mechanism.
ANR-V is more expensive in our video-adapted implementation because LAL introduces thresholding and renormalization overhead in dense coordinate-wise reconstruction.
CoANeRV occupies an intermediate point: it retains attention-based coordinate decoding, but with considerably lower compute than ANR-V and much lower memory at high resolution. This is consistent with the design goal of CoANeRV as a shared-decoder representation model rather than a pure codec or pure lightweight generator.

\paragraph{Memory scaling across videos.}
The relevant deployment-level comparison is not only the per-frame FLOPs, but also how storage grows across a collection of videos. By section~\ref{app:persistent}, per-video decoders scale as $\mathcal{O}(Mp)$, whereas CoANeRV scales as $\mathcal{O}(p+MNd)$. This is exactly the regime in which persistent weights become attractive: when many videos share a common decoder but differ in compact token state.

\paragraph{Training and inference trade-off.}
The shared-decoder design accelerates encoding and reduces multi-video storage, but it does not make coordinate-wise decoding free. In particular, pixel-level attention decoding is slower than convolutional frame generation in some compression settings. We therefore view CoANeRV as optimizing for a broader representation trade-off, not for maximum decode throughput alone.

\subsection{Hyperparameter Sensitivity Analysis}
\label{app:hparam}

We analyze four key hyperparameters: the frequency scale $\sigma$, encoder depth, decoder MLP depth, and batch size. The trends are consistent across datasets and support the chosen operating point.

\begin{figure}[t!]
    \centering
    \includegraphics[width=0.98\linewidth]{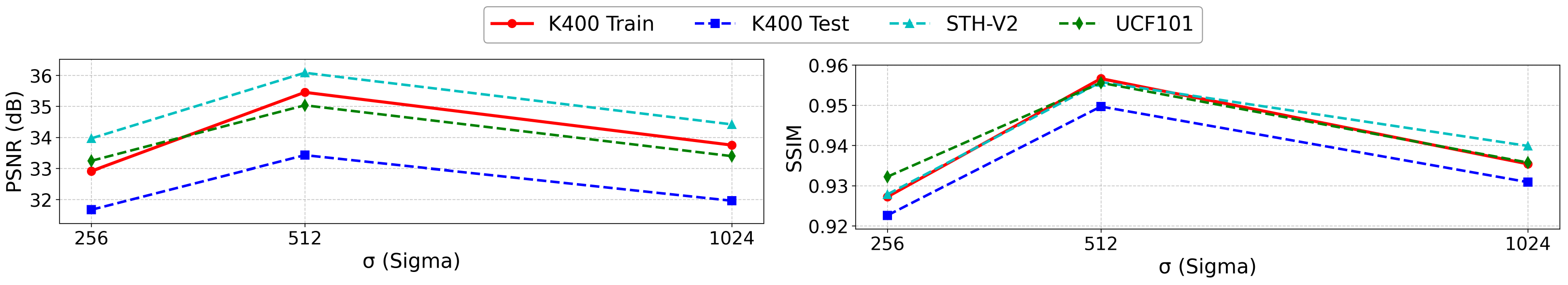}
    \caption{\textbf{Sigma parameter.} Reconstruction performance across training and test sets under different $\sigma$ values.}
    \label{fig:sigma}
\end{figure}

\paragraph{Sigma parameter.}
Figure~\ref{fig:sigma} exhibits an inverted-U trend. A small $\sigma$ restricts frequency coverage and underfits fine spatial structure; an excessively large $\sigma$ overemphasizes high frequencies and hurts generalization. The setting $\sigma=512$ offers the best balance.

\begin{figure}[t!]
    \centering
    \includegraphics[width=0.98\linewidth]{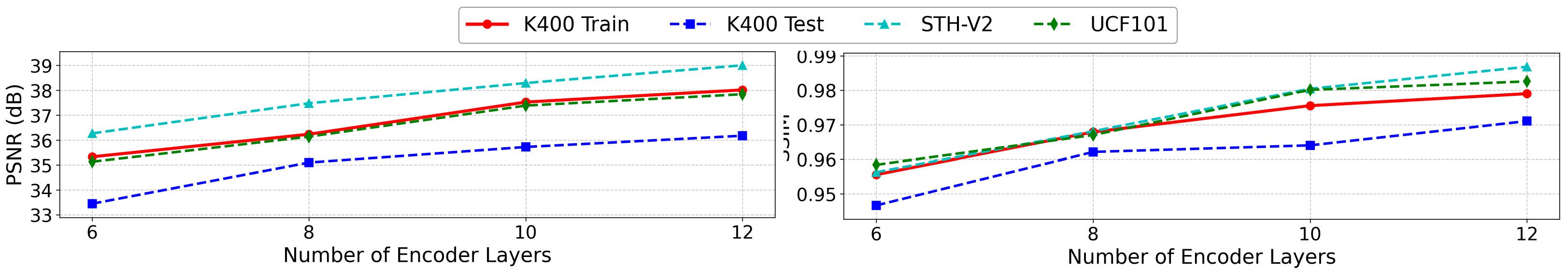}
    \caption{\textbf{Encoder depth.} Performance improves as the encoder depth increases.}
    \label{fig:layers}
\end{figure}

\paragraph{Encoder depth.}
Figure~\ref{fig:layers} shows monotonic gains as the encoder depth increases, indicating that deeper token compression benefits hierarchical spatiotemporal modeling.

\begin{figure}[t!]
    \centering
    \includegraphics[width=0.98\linewidth]{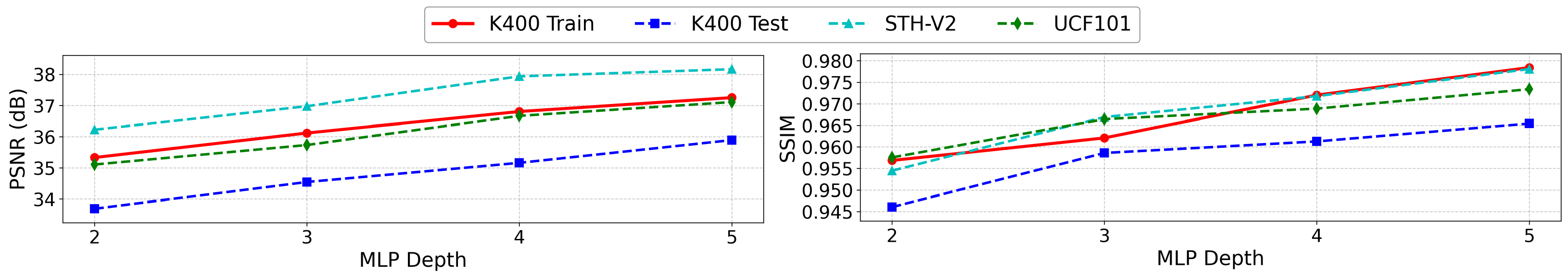}
    \caption{\textbf{MLP depth.} A deeper decoder improves the non-linear coordinate-to-signal mapping.}
    \label{fig:depth}
\end{figure}

\paragraph{MLP depth.}
Figure~\ref{fig:depth} shows that a deeper decoder MLP improves the coordinate-to-signal mapping, with depth $5$ providing the strongest results.


\subsection{Cross-Domain Reconstruction}
\label{app:medical}

We evaluate cross-domain amortized reconstruction without dataset-specific fine-tuning on two echocardiography datasets using models trained on K400 only. The evaluation covers EchoCP~\cite{wang2021echocp} and EchoNet-LVH~\cite{duffy2022high}.

\begin{figure*}[t!]
    \centering
    \includegraphics[width=.98\linewidth]{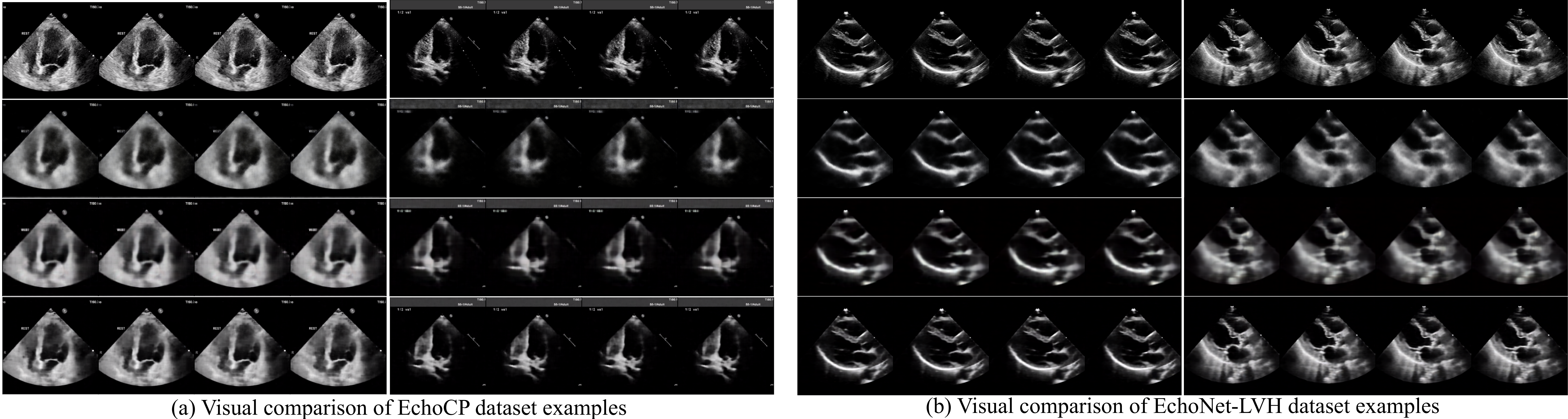}
    \caption{\textbf{Cross-domain qualitative comparison.} Ground truth (top), FastNeRV (second row), ANR-V (third row), and CoANeRV (bottom).}
    \label{fig:echo-visualization}
\end{figure*}

\paragraph{Qualitative results.}
Figure~\ref{fig:echo-visualization} shows that FastNeRV tends to blur anatomical boundaries, ANR-V improves detail recovery but still misses fine structures, and CoANeRV preserves clearer cardiac boundaries and texture under substantial noise.

\begin{table*}[ht]
\centering
\caption{Comparative performance on medical imaging datasets.}
\begin{tabular}{l|c|c|c|c|c}
\hline
\textbf{Dataset} & \textbf{Method} & \textbf{PSNR (dB) $\uparrow$} & \textbf{SSIM $\uparrow$} & \textbf{PSNR Gain} & \textbf{SSIM Gain} \\
\hline
\multirow{3}{*}{EchoCP}
& FastNeRV & 20.53 & 0.607 & - & - \\
& ANR-V & 24.40 & 0.784 & +3.87 & +0.177 \\
& \textbf{CoANeRV} & \textbf{26.85} & \textbf{0.864} & \textbf{+6.32} & \textbf{+0.257} \\
\hline
\multirow{3}{*}{EchoNet-LVH}
& FastNeRV & 22.88 & 0.684 & - & - \\
& ANR-V & 27.45 & 0.835 & +4.57 & +0.151 \\
& \textbf{CoANeRV} & \textbf{29.64} & \textbf{0.916} & \textbf{+6.76} & \textbf{+0.232} \\
\hline
\end{tabular}
\label{tab:medical_reconstruction_comparison}
\end{table*}

\paragraph{Quantitative results.}
Table~\ref{tab:medical_reconstruction_comparison} shows that CoANeRV consistently outperforms FastNeRV and ANR-V on both datasets. The average gain over FastNeRV is 6.54 dB PSNR and 0.245 SSIM, indicating that the token-conditioned coordinate decoder transfers better than the baselines to structurally different video domains.

\subsection{Comparison with Neural Video Compression}
\label{app:nvc}

Although CoANeRV is designed primarily for representation fidelity, we also compare it with neural video codecs DCVC-DC~\cite{li2023neural}, DCVC-FM~\cite{li2024neural}, and DCVC-RT~\cite{jia2025towards}.

\begin{table*}[t]
    \caption{\textbf{Speed analysis.} Encoding / decoding speed (frames per second) across multiple devices.}
    \centering
    \setlength{\tabcolsep}{6pt}
    \begin{tabular}{cc}
        \begin{minipage}{0.47\textwidth}
            \centering
            \resizebox{\linewidth}{!}{
            \begin{tabular}{ l | c | c | c | c}
                \toprule
                Model & A100 & A6000 & 4090 & 2080Ti\\
                \midrule
                DCVC-DC & 3.3 / 4.3 & 1.7 / 2.2 & 2.3 / 2.9 & 0.8 / 1.4 \\
                \midrule
                DCVC-FM & 5.0 / 5.9 & 3.1 / 3.8 & 3.7 / 4.4 & 1.9 / 2.3 \\
                \midrule
                DCVC-RT & 125.2 / 112.8 & 70.4 / 63.8 & 118.8 / 105.3 & 39.5 / 34.1 \\
                \midrule
                CoANeRV & 247.0 / 7.7 & 168.4 / 6.5 & 236.8 / 7.4 & 94.6 / 3.4 \\
                \bottomrule
            \end{tabular}}
            \subcaption{Coding speed on $1920\times1080$ videos.}
        \end{minipage}
        &
        \begin{minipage}{0.47\textwidth}
            \centering
            \resizebox{\linewidth}{!}{
            \begin{tabular}{ l | c | c | c | c}
                \toprule
                Model & A100 & A6000 & 4090 & 2080Ti\\
                \midrule
                DCVC-DC & 6.5 / 7.9 & 3.5 / 4.3 & 5.5 / 6.7 & 2.1 / 2.9 \\
                \midrule
                DCVC-FM & 8.5 / 9.4 & 5.9 / 6.6 & 9.3 / 10.4 & 4.0 / 4.7 \\
                \midrule
                DCVC-RT & 173.9 / 149.2 & 147.3 / 132.5 & 225.1 / 185.2 & 73.3 / 67.0 \\
                \midrule
                CoANeRV & 250.0 / 12.7 & 202.3 / 8.5 & 240.6 / 12.4 & 98.5 / 5.4 \\
                \bottomrule
            \end{tabular}}
            \subcaption{Coding speed on $1280\times720$ videos.}
        \end{minipage}
    \end{tabular}
    \label{tab:compare_complexity}
\end{table*}

\textbf{Encoding Speed.}
Under this representation-oriented protocol, CoANeRV shows high token-extraction throughput because the timed path contains a feed-forward token encoder but excludes quantization, post-hoc Huffman bitstream construction, file I/O, and motion-compensated predictive coding.
This number should not be interpreted as full codec encoding speed, since neural codecs include bitstream generation and optimize a different objective.
On 1080p videos, CoANeRV reports 247.0 fps token-encoding throughput on A100, compared with 125.2 fps for DCVC-RT and 3.3 fps for DCVC-DC.
This difference mainly reflects the exclusion of entropy coding and motion-compensated predictive coding from the timed token-extraction path.

\textbf{Decoding Speed.} CoANeRV decodes at 7.7 fps on A100 for 1080p, lower than DCVC-RT's 112.8 fps. This difference reflects architectural trade-offs: DCVC-RT uses optimized convolutional decoders for parallel frame generation, while CoANeRV performs cross-attention for pixel-level reconstruction.
This makes the current implementation more suitable for offline or representation-oriented settings than for real-time decoding.

\begin{figure}[t!]
    \centering
    \includegraphics[width=0.55\textwidth]{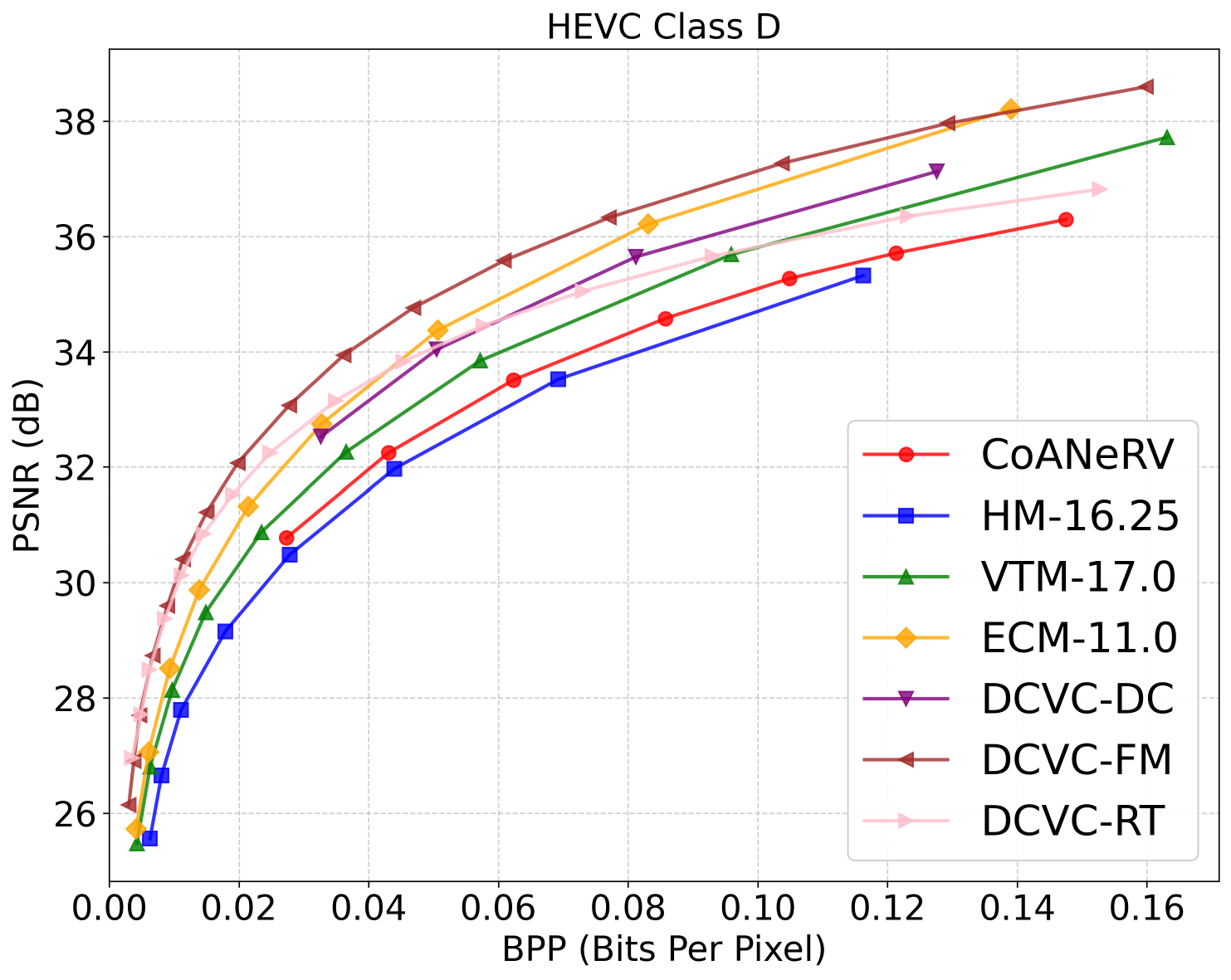}
    \caption{Rate--distortion curves on HEVC Class D sequences.}
    \label{fig:Full-RD-Curve}
\end{figure}

\paragraph{Rate--distortion.}
Figure~\ref{fig:Full-RD-Curve} provides an auxiliary rate--distortion visualization on HEVC Class D sequences.
The result should be interpreted as a representation-storage comparison rather than a full codec benchmark: CoANeRV has no learned entropy model or end-to-end rate-control mechanism, and the optional post-hoc Huffman code used for storage accounting is not included in the timed token-extraction path.

\subsection{High-Resolution Scalability Analysis}
\label{app:highres}

We compare CoANeRV with ANR-V at $512\times512$ and $1024\times1024$ resolutions. The results are summarized in Table~\ref{tab:anr_res}.

\begin{table*}[t!]
\centering
\caption{\textbf{CoANeRV vs. ANR-V with different resolutions.} CoANeRV achieves better quality with much lower memory and training cost. Memory is the recorded peak GPU footprint for one minibatch in GB, using $1\,\mathrm{GB}=1024\,\mathrm{MiB}$.}
\resizebox{.98\textwidth}{!}{%
\begin{tabular}{@{}lc|cc|ccc|cccc|cccc}
\toprule
\multirow{2}{*}{Methods} & \multirow{2}{*}{F} & \multirow{2}{*}{\makecell{Frame \\ Resolution}} & \multirow{2}{*}{$\#\hat\theta'$} & \multirow{2}{*}{Epoch} & \multirow{2}{*}{\makecell{Peak memory \\ / batch (GB) $\downarrow$}} & \multirow{2}{*}{\makecell{GPU \\ hrs $\downarrow$}} & \multicolumn{4}{c}{PSNR $\uparrow$} & \multicolumn{4}{c}{SSIM $\uparrow$} \\
 &  &  &  &  &  &  & Train & K400 & SthV2 & UCF101 & Train & K400 & SthV2 & UCF101 \\
\midrule
ANR-V-S & 4 & 512  & 27k & 50 & 64 & 275 & 27.9 & 27.2 & 28.0 & 27.0 & 0.833 & 0.819 & 0.834 & 0.829 \\
CoANeRV-S & 4 & 512  & 27k & 50 & 5.2 & 62 & \textbf{29.3} & \textbf{29.5} & \textbf{29.0} & \textbf{29.8} & \textbf{0.855} & \textbf{0.846} & \textbf{0.856} & \textbf{0.855} \\
ANR-V-M & 4 & 512  & 36k & 50 & 76 & 285 & 28.2 & 27.8 & 28.1 & 28.2 & 0.842 & 0.830 & 0.842 & 0.839 \\
CoANeRV-M & 4 & 512  & 36k & 50 & 5.3 & 66 & \textbf{29.8} & \textbf{29.7} & \textbf{29.6} & \textbf{29.9} & \textbf{0.860} & \textbf{0.852} & \textbf{0.860} & \textbf{0.860} \\
ANR-V-L & 4 & 512  & 54k & OOM & OOM & OOM & OOM & OOM & OOM & OOM & OOM & OOM & OOM & OOM \\
CoANeRV-L & 4 & 512  & 54k & 50 & 5.3 & 70 & \textbf{30.7} & \textbf{30.1} & \textbf{30.9} & \textbf{30.5} & \textbf{0.870} & \textbf{0.865} & \textbf{0.872} & \textbf{0.873} \\
\midrule
ANR-V-S & 1 & 1024  & 27k & 50 & 64 & 278 & 26.9 & 27.3 & 26.8 & 26.3 & 0.821 & 0.801 & 0.814 & 0.807 \\
CoANeRV-S & 1 & 1024  & 27k & 50 & 5.2 & 64 & \textbf{29.1} & \textbf{29.1} & \textbf{28.5} & \textbf{28.8} & \textbf{0.839} & \textbf{0.827} & \textbf{0.816} & \textbf{0.813} \\
ANR-V-M & 1 & 1024  & 36k & 50 & 76 & 305 & 27.4 & 28.4 & 27.9 & 27.2 & 0.824 & 0.801 & 0.813 & 0.808 \\
CoANeRV-M & 1 & 1024  & 36k & 50 & 5.3 & 68 & \textbf{29.6} & \textbf{28.2} & \textbf{29.9} & \textbf{29.7} & \textbf{0.843} & \textbf{0.831} & \textbf{0.839} & \textbf{0.843} \\
ANR-V-L & 1 & 1024  & 54k & OOM & OOM & OOM & OOM & OOM & OOM & OOM & OOM & OOM & OOM & OOM \\
CoANeRV-L & 1 & 1024  & 54k & 50 & 5.3 & 71 & \textbf{30.2} & \textbf{29.5} & \textbf{30.4} & \textbf{30.2} & \textbf{0.854} & \textbf{0.840} & \textbf{0.852} & \textbf{0.856} \\
\bottomrule
\end{tabular}}
\label{tab:anr_res}
\end{table*}

\paragraph{Resolution scaling.}
At $512$ resolution with four frames, CoANeRV already outperforms ANR-V at matched model sizes, and the gap grows as model capacity increases. At $1024$ resolution, CoANeRV continues to run in 5.2--5.3 GB per batch, whereas ANR-V requires 64--76 GB and fails for large models. These results are consistent with section~\ref{app:block_query}: the block-query decoder preserves the reconstruction rule while reducing peak attention memory.

\paragraph{Encoding efficiency.}
Figure~\ref{fig:encoding} compares CoANeRV with gradient-based NeRV optimization.
The empirical $10^5\times$ speedup is consistent with the asymptotic argument in Section~\ref{app:persistent}: CoANeRV performs one feed-forward encoding pass instead of repeated per-video optimization steps.

\subsection{Token Quantization for Storage Efficiency}
\label{sec:token_quant}

To reduce per-video storage overhead, we quantize the video tokens $\mathbf{T}_v$ while keeping the shared decoder in full precision. Each token element is quantized by
\begin{equation}
\hat{\mathbf{T}}_v =
\mathrm{Round}\!\left(\frac{\mathbf{T}_v-z}{s}\right)s + z,
\label{eq:quant}
\end{equation}
where
\(z=(\mathbf{T}_v)_{\min}\) and
\(s=((\mathbf{T}_v)_{\max}-(\mathbf{T}_v)_{\min})/(2^b-1)\) for \(b\)-bit quantization.

\begin{table}[h!]
\centering
\caption{Ablation study on token quantization. Decoder weights remain in 32-bit floating point.}
\label{tab:supple-weight_token_ablation}
\begin{tabular}{c|ccc|ccc}
\toprule
\multirow{2}{*}{Bits} & \multicolumn{3}{c|}{PSNR $\uparrow$} & \multicolumn{3}{c}{SSIM $\uparrow$} \\
 & K400 & SthV2 & UCF101 & K400 & SthV2 & UCF101 \\
\midrule
32 (Full) & 33.02 & 33.33 & 33.52 & 0.938 & 0.942 & 0.947 \\
\midrule
8 & 32.99 & 33.30 & 33.49 & 0.938 & 0.942 & 0.947 \\
7 & 32.89 & 33.20 & 33.39 & 0.937 & 0.941 & 0.946 \\
6 & 32.51 & 32.82 & 33.01 & 0.933 & 0.937 & 0.942 \\
5 & 31.20 & 31.51 & 31.70 & 0.919 & 0.923 & 0.928 \\
4 & 27.77 & 28.08 & 28.27 & 0.866 & 0.870 & 0.875 \\
\bottomrule
\end{tabular}
\end{table}

The results show that token-space storage admits substantial post-training compression: 8-bit quantization changes K400 PSNR only from 33.02 to 32.99 dB, while 6-bit tokens retain 32.51 dB/0.933 SSIM. Because the decoder is shared and fixed, this storage analysis can focus on the per-video state. It does not include quantization-aware training (QAT), a learned entropy model, side-information accounting for such a model, or end-to-end rate control; these are future codec-oriented extensions rather than claims of the present representation study.

\begin{table}[ht]
\centering
\caption{Complete Hyperparameter Settings}
\begin{tabular}{l|l}
\hline
\textbf{Parameter} & \textbf{Value} \\
\hline
\textbf{CoANeRV Architecture} & \\
Video-token dimension ($d$) & 72 \\
Video-token length ($N$) & 384 \\
Patch-token self-encoder depth & 0 \\
Number of token-former layers ($L$) & 6 \\
Token-former width ($D$) & 768 \\
Token-former attention heads & 6 \\
Token-former per-head dimension & 64 \\
Token-former feed-forward dimension & 3072 \\
Spatial tile side length ($s$) & 64 \\
Queries per tile ($M=Ts^2$) & $T\cdot64^2$ \\
Temperature parameter ($\tau$) & 0.4 \\
\hline
\textbf{Shared Coordinate Decoder Parameters} & \\
Coordinate input dimension & 3 \\
Signal output channels & 3 \\
Output bias & 0.5 \\
Network depth & 2 \\
Hidden dimension & 72 \\
Number of attention heads & 6 \\
Per-head attention dimension \(d_h\) & 64 \\
PE dimension & 24 \\
PE sigma & 512 \\
Activation function & SiLU \\
Rescale & No \\
\hline
\textbf{Training Parameters} & \\
Optimizer & AdamW \\
Learning rate & $10^{-4}$ \\
Weight decay & $10^{-2}$ (PyTorch default) \\
Batch size & 2 \\
Total epochs & 150 \\
Warmup epochs & 0 \\
LR schedule & $1\times$ through epoch 135; $0.1\times$ for epochs 136--150 \\
\hline
\textbf{Data Processing} & \\
Input resolution & $256\times256$ \\
Frame sampling & First $F$ consecutive frames \\
Color space & YUV \\
Normalization & $[0,1]$ \\
\hline
\end{tabular}
\label{tab:param}
\end{table}

\subsection{Detailed Training Configuration}
\label{app:training}

Table~\ref{tab:param} summarizes the full architecture and optimization configuration. All models are implemented in PyTorch~\cite{paszke2019pytorch} and trained with AdamW~\cite{loshchilov2017decoupled}. We use SiLU activations~\cite{elfwing2018sigmoid} and train on 1 NVIDIA A800 GPU with Intel Xeon Gold 6430 CPUs. For each video, the loader selects the first $F$ consecutive decoded frames; each selected frame is then resized and center-cropped to $256\times256$, converted to YUV, and normalized to $[0,1]$.

\subsection{Dataset Splits and Filtering}
\label{app:dataset_manifest}

Each dataset split is defined by a fixed list of video identifiers. Videos with fewer than the requested $F$ decoded frames are excluded deterministically. Table~\ref{tab:dataset_manifest} reports the split sizes before and after filtering for the main 4-, 8-, and 16-frame protocols. Thus, ``10K K400 training videos'' denotes the 10,000-video split before minimum-frame filtering; 9,997 videos remain in the 16-frame setting. The training subset was not constructed using class-aware balancing.

\begin{table}[t]
\centering
\small
\caption{Dataset split sizes before and after deterministic minimum-frame filtering.}
\label{tab:dataset_manifest}
\begin{tabular}{lrrrr}
\toprule
Dataset split & Before filtering & 4F & 8F & 16F \\
\midrule
K400 train & 10,000 & 10,000 & 10,000 & 9,997 \\
K400 validation & 13,732 & 13,732 & 13,732 & 13,731 \\
Something-Something V2 evaluation & 20,000 & 20,000 & 20,000 & 19,947 \\
UCF101 evaluation & 3,500 & 3,500 & 3,500 & 3,500 \\
\bottomrule
\end{tabular}
\end{table}

The exact video identifiers for each split are provided with the released code, enabling reconstruction of the evaluated subsets without redistributing any video data. Dataset access remains subject to the original licenses and terms of use.

\paragraph{Broader Impact.}
CoANeRV may benefit video storage, transmission, and scientific or medical video analysis by improving reconstruction quality and reducing encoding overhead. Potential risks include more efficient storage or reconstruction of privacy-sensitive visual data. Since the method reconstructs input videos rather than generating new semantic content, misuse risks are lower than those of unconstrained video generation models, but deployment should respect dataset consent, privacy, and access-control requirements.

\paragraph{Datasets and Assets.}
We use publicly available benchmark datasets, including Kinetics-400, Something-Something V2, UCF101, HEVC test sequences, EchoCP, and EchoNet-LVH, and cite their original sources. All datasets are used only for research evaluation under their respective licenses or terms of use. Baseline implementations and software dependencies, including PyTorch, are credited through the corresponding citations where applicable.

\end{document}